\documentclass{article}

\usepackage{PRIMEarxiv}

\usepackage[utf8]{inputenc} 
\usepackage[T1]{fontenc}    
\usepackage{hyperref}       
\usepackage{url}            
\usepackage{booktabs}       
\usepackage{amsfonts}       
\usepackage{nicefrac}       
\usepackage{microtype}      
\usepackage{lipsum}
\usepackage{fancyhdr}       
\usepackage{graphicx}       
\graphicspath{{media/}}     

\usepackage{bm}
\usepackage{amsmath}
\usepackage{hhline}

\title{Examining the Proximity of Adversarial Examples to Class Manifolds in Deep Networks
}

\author{
  \v{S}tefan P\'oco\v{s}, Iveta Be\v{c}kov\'a, Igor Farka\v{s} \\
  Faculty of Mathematics, Physics and Informatics \\
  Comenius University in Bratislava, Slovak Republic \\
  \texttt{\{stefan.pocos,iveta.beckova,igor.farkas\}@fmph.uniba.sk} \\
  }
  
\begin{document}
\maketitle

\begin{abstract}
Deep neural networks achieve remarkable performance in multiple fields. However, after proper training they suffer from an inherent vulnerability against adversarial examples (AEs). In this work we shed light on inner representations of the AEs by analysing their activations on the hidden layers. We test various types of AEs, each crafted using a specific norm constraint, which affects their visual appearance and eventually their behavior in the trained networks. Our results in image classification tasks (MNIST and CIFAR-10) reveal qualitative differences between the individual types of AEs, when comparing their proximity to the class-specific manifolds on the inner representations. We propose two methods that can be used to compare the distances to class-specific manifolds, regardless of the changing dimensions throughout the network. Using these methods, we consistently confirm that some of the adversarials do not necessarily leave the proximity of the manifold of the correct class, not even in the last hidden layer of the neural network. Next, using UMAP visualisation technique, we project the class activations to 2D space. The results indicate that the activations of the individual AEs are entangled with the activations of the test set. This, however, does not hold for a group of crafted inputs called the rubbish class. We also confirm the entanglement of adversarials with the test set numerically using the soft nearest neighbour loss.
\end{abstract}


\section{Introduction}
Studies of the past years have shown that  carefully crafted minor perturbations can be added to an input to alter the predictions. Such modified inputs are called adversarial examples (AEs) \cite{szegedy_int}. Their existence poses a serious security risk not only in the domain of image classification, but also for malware detection \cite{malware}, automatic speech recognition \cite{carlini_audio} and many more applications. For that reason, the study of robustness became a hot topic. 
Earlier experiments \cite{tsipras2019robustness} suggested that obtaining networks, which are both accurate and robust is not possible. However, some studies argue that robust and accurate networks are indeed achievable \cite{Stutz2019,gilmer2018adversarial}. They claim that defence against AEs is basically a good generalisation, so robustness and accuracy may not necessarily be contradicting goals. 
Numerous defence mechanisms have been proposed, but most of them yield only seemingly better results, due to the gradient obfuscation \cite{obfuscated}. So far, the most promising one is adversarial training \cite{madry_towards} (based on enhancing the training set with adversarial inputs), but it is computationally demanding, it cannot guarantee absolute robustness and the defender is bounded to a specific attack type.

In our work we focus on adversarial examples from the perspective of data manifolds, formed by the hidden-layer activations of the training set, test set, the set of generated AEs and the rubbish class examples (Section 2). 

We conduct a computational analysis, where we compare the average distances of AEs vs original images to different classes (Section 3). To eliminate the issues when comparing distances in spaces with different dimensions, we propose two methods of assessing and comparing the class correspondences of AEs in the hidden-layer representations and show consistent distinctions between inner representations of different types of AEs and rubbish class examples (Section 4). Next, we show that AEs are entangled with the test set and their entanglement gradually decreases progressing through the network (Section 5). We end by concluding remarks (Section 6).
Overall, these results provide novel qualitative insights into the geometry of AEs.

\section{Experimental setup}

\subsection{Models}
For evaluations of adversarial examples and their behaviour during the classification we use MNIST and CIFAR-10 datasets
of which the latter is a somewhat more difficult task for classification. To observe the behaviour of AEs, we train two different types of networks on MNIST: a fully connected feed-forward network with two hidden layers, each containing 128 neurons followed by ReLU, and a convolutional network with two convolutional layers, each containing 16 filters followed by  max-pooling and ReLU activation function.

In order to classify CIFAR-10 images, we use a deep convolutional network, consisting of three VGG-type \cite{simonyan2015deep} blocks, after which an additional fully connected layer of 256 neurons is applied (before the output layer). In the case of CIFAR-10, we also use dropout \cite{hinton2012improving} and batch normalisation \cite{ioffe2015batch}. 

To eliminate the effects of additional factors on internal representations and behaviours of the networks, we deliberately do not work with the state-of-the-art networks for classifying the datasets. Although, we achieve satisfying average accuracies for our purposes, i.e. $98.0$ \%  using a fully connected network, $98.9$ \% for the convolutional network, both trained on MNIST using SGD, and $87.2$ \% for the network trained on CIFAR-10 using Adam optimizer \cite{Adam}.

\subsection{Attack methods}
The quality of defences against AEs is rising and so is the number of attacks. Therefore, nowadays there already exist a lot of distinct methods for generating AEs, each having its pros and cons. In this section we describe the attacks we chose for our experiments, their key features and the differences between them.

Usually, a method for generating AEs is dependent on a set of hyperparameters, such as $L_p$ norm used to measure the perturbation, the number of iterations, confidence of resulting adversarial examples, and the perturbation magnitude. However, when a defence method is designed, or some kind of analysis is performed, the authors frequently select a specific attack algorithm, fix certain hyperparameters and carry out an experiment. Afterwards, a question arises whether the results are robust enough for qualitatively different types of AEs. In hope of mitigating the issues with a too specific attack, we generate AEs to be as distinct as possible and we analyse their behaviour inside deep networks.

When considering the perturbation magnitude, the selection of a distance metric is crucial. A common choice is $L_p$ metric. Surprisingly, the resulting AEs for different $p$ are not universal and generating the examples using different $L_p$ norm usually yields a completely different kind of AEs.
To ensure the diversity of studied AEs, we use the following four types of attacks, optimized for a specific norm constraint:\footnote{Rubbish class examples (also called fooling examples or false positives) do not meet the definition of AEs, however they also provide useful insights into robustness.}

\begin{itemize}
\item \textbf{{$\boldsymbol{L_{\infty}}$ constraint:}} We utilise projected gradient descent (PGD), which was introduced in \cite{madry_towards}. Given a perturbation magnitude $\epsilon$, PGD finds the most fooling image with perturbation no greater than $\epsilon$. The attack usually finds a perturbation of magnitude close to $\epsilon$, so in order to have various perturbations, we generate AEs for MNIST $\epsilon \in \{0.01, 0.02, ..., 0.15\}$ and for CIFAR-10 $\epsilon \in \{0.01, 0.02, ..., 0.05\}$.
\item \textbf{{$\boldsymbol{L_2}$ constraint:}} To generate AEs with a low $L_2$ perturbation, we use Carlini \& Wagner (CW) attack \cite{carlini_cw} which belongs to the state-of-the-art methods for generating AEs with minimal perturbations. The optimization procedure is aimed at creating an AE with the smallest possible perturbation $\boldsymbol\delta$ according to:
\begin{align}
\begin{split}
    &\text{minimize} \quad c \cdot f(\mathbf{x}+\boldsymbol\delta) + \|\boldsymbol\delta\|_2 , \\
    &\text{subject to} \quad \mathbf{x} + \boldsymbol\delta \in [0,1]^n.
\end{split}
\label{eq:CW}
\end{align}

\item \textbf{{$\boldsymbol{L_1}$ constraint:}} In \cite{chen_ead} it was argued that $L_1$ norm accounts for the total variation of perturbations, and in the previous studies little effort was dedicated to crafting AEs minimizing $L_1$ norm. Therefore, the authors modified the CW algorithm by adding a term penalizing $L_1$ norm of the perturbation, possibly even completely eliminating the $L_2$ term, resulting in somewhat different optimization procedure. Therefore, we use this attack to generate another set of AEs. 
\item \textbf{{$\boldsymbol{L_0}$ constraint:}} In this attack the goal is to change the smallest number of pixels, in order to achieve misclassification. A popular choice for this constraint is a semi-black box attack based on differential evolution, introduced in \cite{su_one_pixel}. However, when dealing with networks with available gradients, a white-box attack can be more effective, providing faster and qualitatively comparable results. For that reason, we propose $L_0$ attack as follows: First, we calculate the gradient of loss w.r.t. the input image and choose a pixel with the greatest absolute gradient. Second, we perform a grid search for pixel values and choose the one, which yields the maximal decrease of the output probability for the correct class. This procedure is then repeated (perturbing one pixel at a time), until misclassification occurs, or the stopping criterion is met (perturbation of 50 pixels).
\end{itemize}

In our studies we also include \textbf{rubbish class} (RC) examples \cite{goodfellow_explaining} e.g. seemingly random noise patterns which are classified by the network as one of the target classes with high confidence, in our case at least $95$ \%. We generate RC examples in two distinct ways. The first is based on sampling from a uniform distribution for each pixel value and using PGD to find an AE with a small perturbation, so the noise pattern does not change a lot.
The second method differs only in the initialisation, where the color value of each pixel is sampled from the probability distribution function of the pixels in the same position -- we use a random selection of a pixel from the training data. In the following text, we denote these groups as ${\rm RC}_{rnd}$ and ${\rm RC}_{distrib}$, respectively.

We generate a body of AEs for each attack type, $\approx$12,000 AEs for each of the three used networks.\footnote{We use ART (Adversarial Robustness Toolbox) \cite{art2018} for all attacks except $L_0$.} In most of the cases ($L_0$, $L_1$, $L_2$, $L_{\infty}$) we use a non-targeted attack i.e.~it does not matter what kind of misclassification occurs as long as the adversarial image is not classified correctly. However, to support the diversity of the generated inputs, for the rubbish class we use a targeted attack (in case of PGD this corresponds to maximizing the output probability of the target class instead of minimizing the output of the correct class) and we set the target step by step to all classes. Using a targeted attack for the analysis of rubbish class examples is favourable, because without that the networks tend to classify a big margin of rubbish class inputs as belonging to the same class (for MNIST as '8' and for CIFAR-10 as 'frog').

\begin{figure*}[t!]
\centering
\includegraphics[width=0.49\columnwidth]{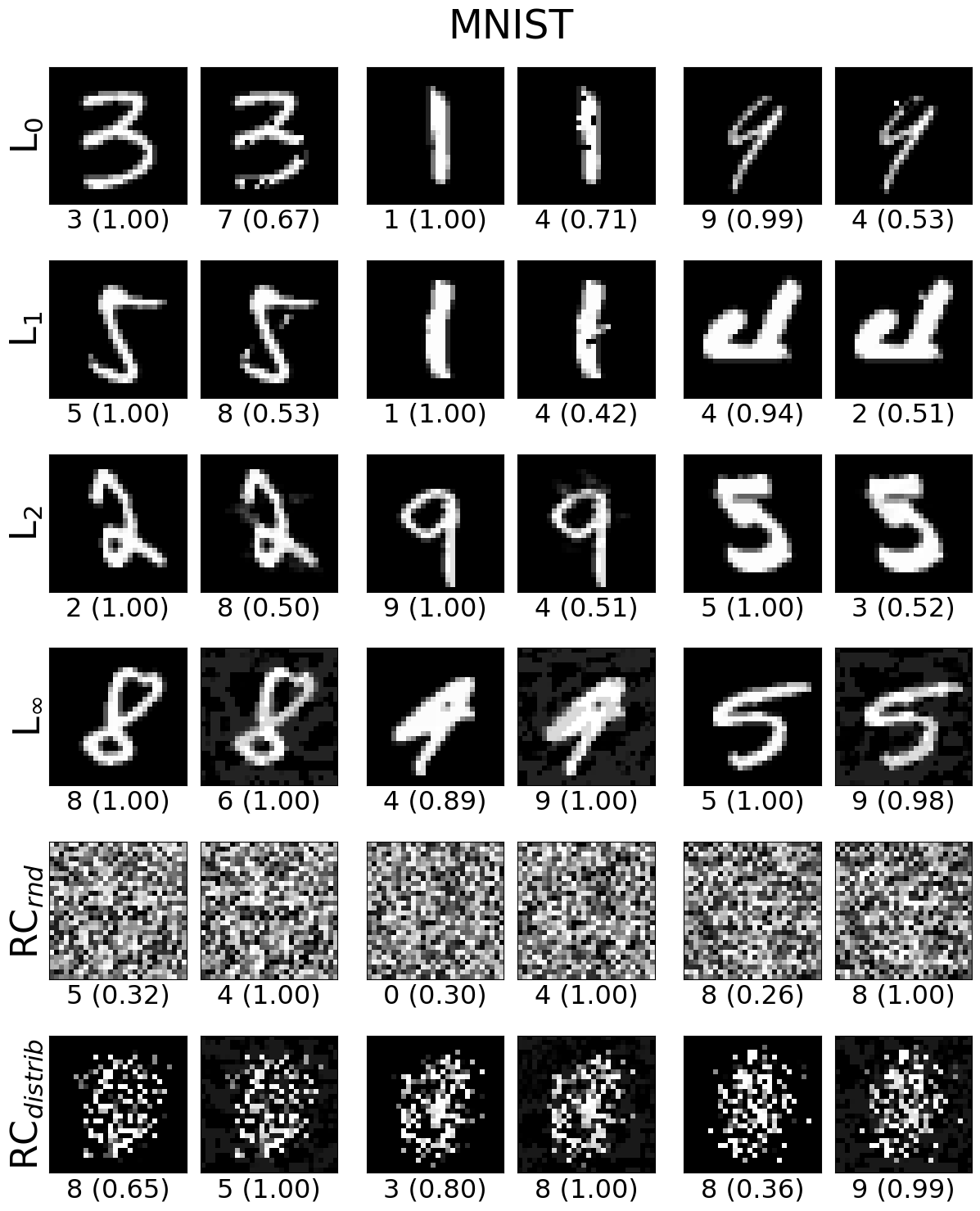}
\includegraphics[width=0.481\columnwidth]{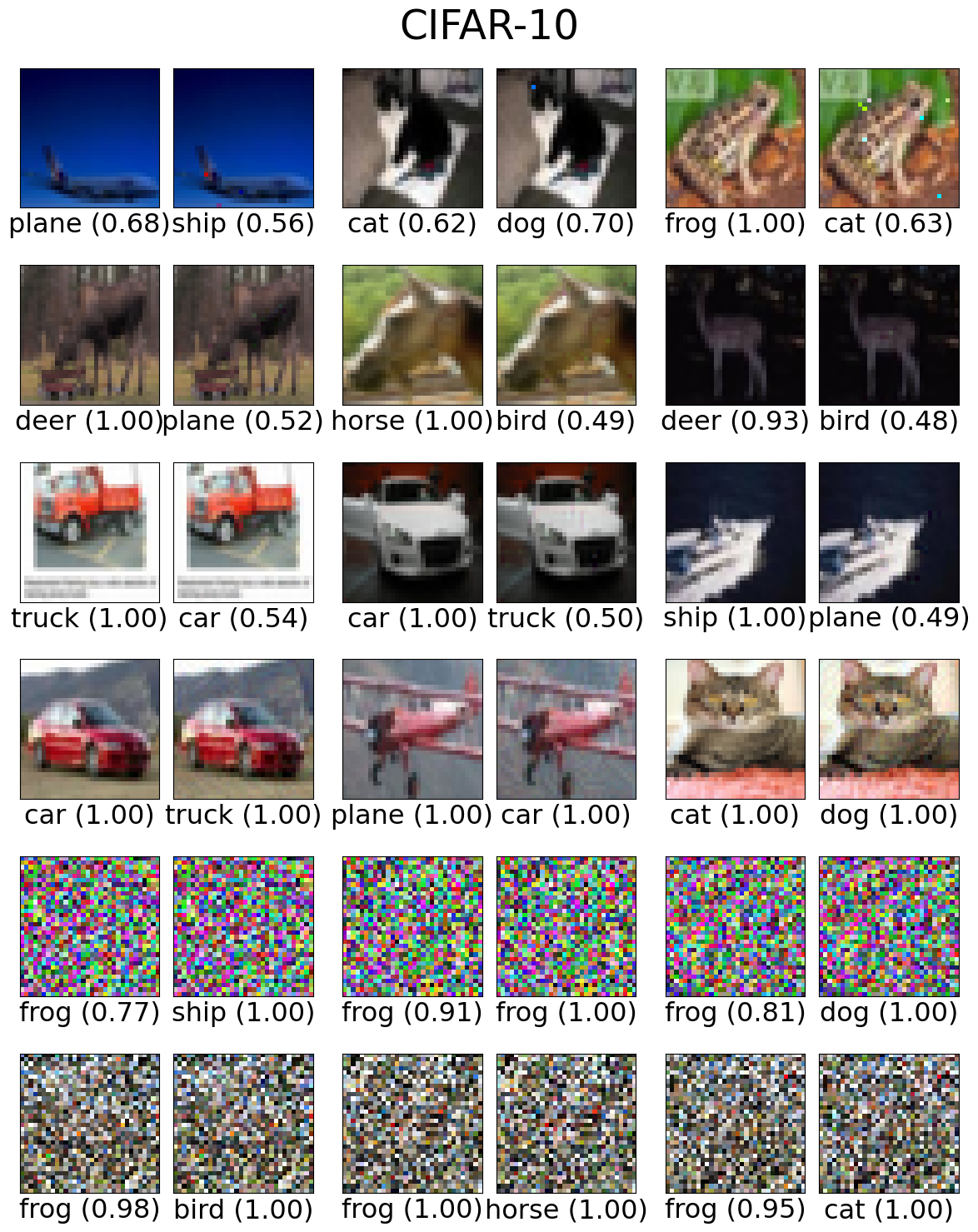}
\caption{An illustration of all 6 types of attacks crafted for convolutional networks on MNIST and CIFAR-10. For each attack, three pairs of original (left) vs. adversarial (right) images are shown with their corresponding predicted class (and confidence).}
\label{fig:AE}
\end{figure*}

A random sample of the generated AEs for the two used datasets is shown in Fig.~\ref{fig:AE}. The differences between AEs generated by the mentioned methods are easy to see by a naked eye, and they can also be easily demonstrated using distances of AEs to their original images in individual norms (shown in Table \ref{tab:distances}).

\begin{table}
    \centering
    \small
    \caption{Average distances of the AEs crafted using different attacks. We can see that the individual attacks indeed minimize the distance of an AE to the original input in the norm, which was chosen as a constraint for them.}
    \tabcolsep=0.08cm
    \begin{tabular}{| c || c  c  c  c || c  c  c  c || c  c  c  c |}
    \hline
    \multicolumn{1}{|c||}{ } & \multicolumn{4}{|c||}{\textbf{MNIST FC}}  & \multicolumn{4}{|c||}{\textbf{MNIST Conv.}} & \multicolumn{4}{|c|}{\textbf{CIFAR-10}} \\
     & $L_0$ & $L_1$ & $L_2$ & $L_{\infty}$ & $L_0$ & $L_1$ & $L_2$ & $L_{\infty}$ & $L_0$ & $L_1$ & $L_2$ & $L_{\infty}$ \\
    \hline
    Our ($L_0$) & $\boldsymbol{10.42}$ & $8.50$ & $2.69$ & $1.00$ & $\boldsymbol{13.14}$ & $9.91$ & $2.86$ & $1.00$ & $\boldsymbol{9.78}$ & $10.01$ & $2.19$ & $0.81$ \\
    EAD($L_1$) & $51.88$ & $\boldsymbol{8.28}$ & $1.68$ & $0.73$ & $42.14$ & $\boldsymbol{7.82}$ & $1.83$ & $0.84$ & $66.34$ & $\boldsymbol{1.97}$ & $0.29$ & $0.11$ \\
    ${\rm CW}(L_2)$& $286.52$ & $13.21$ & $\boldsymbol{1.13}$ & $0.26$ & $180.21$ & $11.83$ & $\boldsymbol{1.38}$ & $0.43$ & $428.98$ & $3.85$ & $\boldsymbol{0.21}$ & $0.04$ \\
    ${\rm PGD}(L_{\infty})$ & $493.27$ & $47.06$ & $2.21$ & $\boldsymbol{0.11}$ & $516.68$ & $49.28$ & $2.35$ & $\boldsymbol{0.12}$ & $1022.81$ & $76.83$ & $1.46$ & $\boldsymbol{0.03}$ \\
    \hline
    \end{tabular}
\label{tab:distances}
\end{table}

\section{Distances to classes}

A typical neural network classifier takes an input (an image in our case) and returns the predicted class label. However, the process of how the input image gets transformed into a certain output class is highly non-linear and more importantly, non-transparent. This is one of the drawbacks of modern machine learning. Not only are the representations hard to interpret, but in addition, the inherent dimensionalities of the spaces (i.e.~hidden layers) are usually too high, preventing the use of multiple methods for their analysis.

In this section we focus on the question why AEs end up yielding wrong output. Is it a continuous process of slowly diverging from the correct class manifold, a sudden leap to the wrong class category in the activation space (at hidden layers), or only wrong generalization caused at the very last layer in the network?

\begin{figure*}[t!]
\centering
\includegraphics[width=0.33\columnwidth]{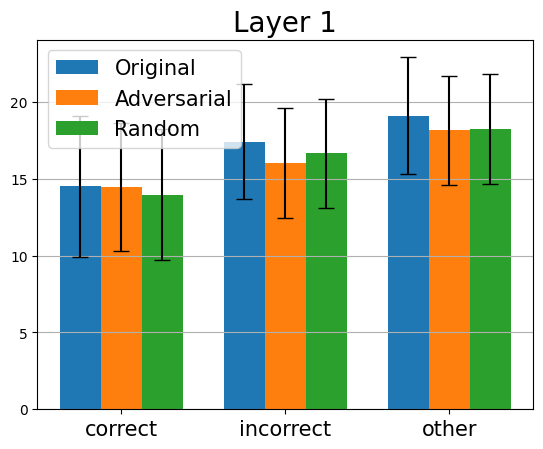}
\includegraphics[width=0.33\columnwidth]{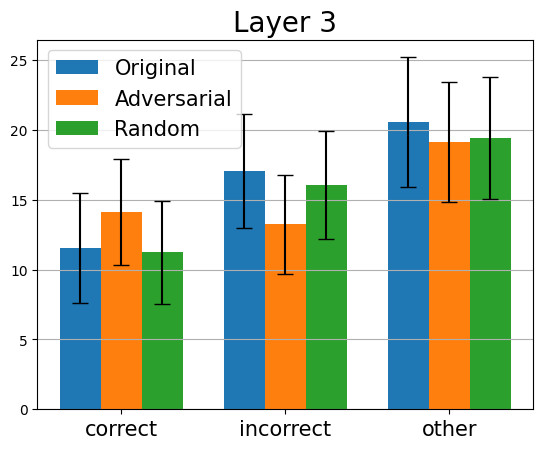}
\includegraphics[width=0.33\columnwidth]{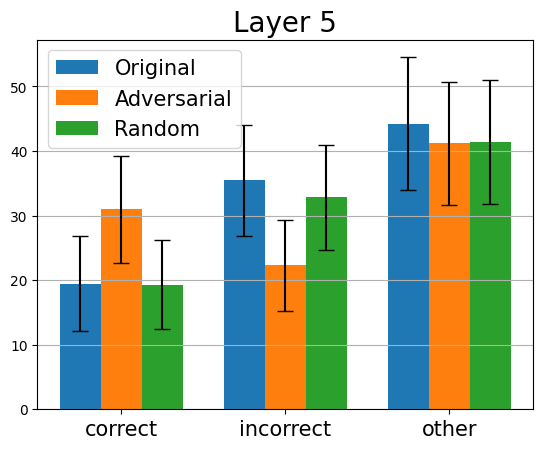}
\caption{Average distances of specific input activation (original image, AE or original modified by random noise) to three categories of data -- the original (correct) class, the incorrect class (found by the attack) and the other classes (random). The data were generated by modifying the MNIST dataset and for generating AEs we used the PGD attack.}
\label{fig:advorigrnd1}
\end{figure*}

\begin{figure*}[t!]
\centering
\includegraphics[width=0.33\columnwidth]{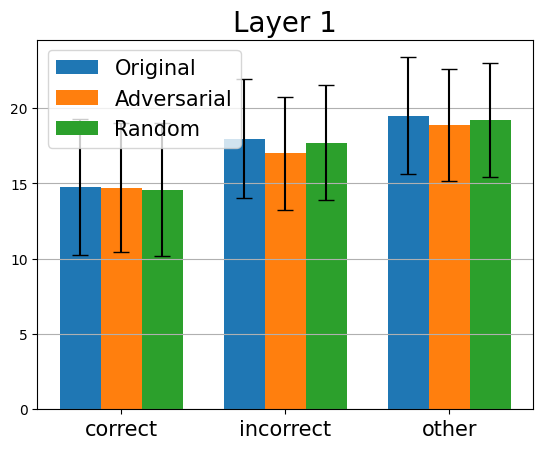}
\includegraphics[width=0.33\columnwidth]{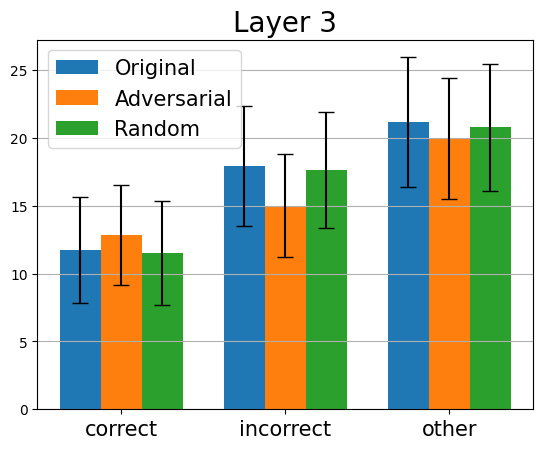}
\includegraphics[width=0.33\columnwidth]{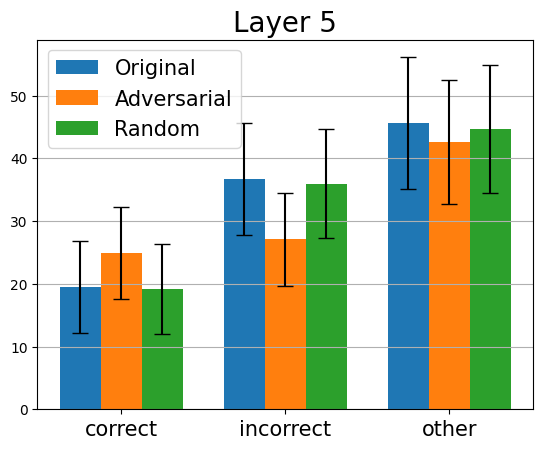}
\caption{Average distances of specific input activation to three categories of data -- the original (correct) class, the class found by the attack (incorrect) and the other (random) classes. The data were generated by modifying the MNIST dataset and for generating the AEs we used the CW method.}
\label{fig:advorigrnd2}
\end{figure*}

We inspect this phenomenon by comparing the distances in the activation spaces. To be more general, we create triples $O_i, A_i, R_i$, where $O_i$ denotes the original image, $A_i$ the adversarial image and $R_i$ represents original image modified by random noise, where the perturbation magnitude is the same as the one used for creating $A_i$. Having these triples, we  systematically calculated the average distances to three different groups of data in the activation space. The first is the distance to the correct class, second is the distance to the incorrectly predicted class, which was found by adversarial attack and the third is the average distance to the activations of the rest of the examples in the training set. Exemplar visualisations are depicted in Figs.~\ref{fig:advorigrnd1} and \ref{fig:advorigrnd2}. 

From these experiments we see that AEs tend to get closer to the incorrect class later in the network and their distance to the original class is slowly rising. We also see that on average, the incorrect class is closer than the rest of the classes.

We note that there is a non-trivial shift of AEs towards the incorrect class, which is not visible when analyzing the original examples. 
Another interesting observation is that on average, the distance of original clean examples to the incorrect class is smaller than to other classes. This might mean that in most cases the adversarial attack causes the output to shift towards the closest incorrect class.

Unfortunately, using statistical comparison methods such as the one described above cannot capture this process of movement from one manifold towards another, due to different dimensionalities of the activation spaces. These comparisons would be inconsistent.

\section{Proximity to manifolds}
\label{sec:proximity}
To illuminate the process of gradual mapping of an adversarial image onto an incorrect output, we propose two methods. Each of them analyses the distances to the manifolds of activations at hidden layers. These methods allow us to compare the distances throughout the network, even though the dimensionalities of the layers (activation spaces) differ.

\subsection{Counting the nearest neighbours}
In this method, we leverage the idea of searching for the nearest neighbours in the space of hidden-layer representations, inspired by \cite{papernot2018deep}.
First, for a given network and a chosen attack, we pick a subset of AEs ($Adv_{C_o \rightarrow C_p}$), which consists of AEs crafted from images belonging to the original class $C_o$ that are misclassified as the predicted class $C_p$, where $p \neq o$. The next step is to find the $k$ nearest neighbours in the activation space for each $\mathbf{x} \in Adv_{C_o \rightarrow C_p}$, where we only consider the activations of those data from the training set, which belong to $C_o$ or $C_p$. Then we calculate the $k_o/k$ ratio, where $k_o$ refers to the number of points from the manifold of the original, correct class. This way, we can visualise the average ratio (considering $\forall \mathbf{x} \in Adv_{C_o \rightarrow C_p}$) and see how this value develops throughout the network.

\begin{figure*}[t!]
\centering
\includegraphics[width=0.9\columnwidth]{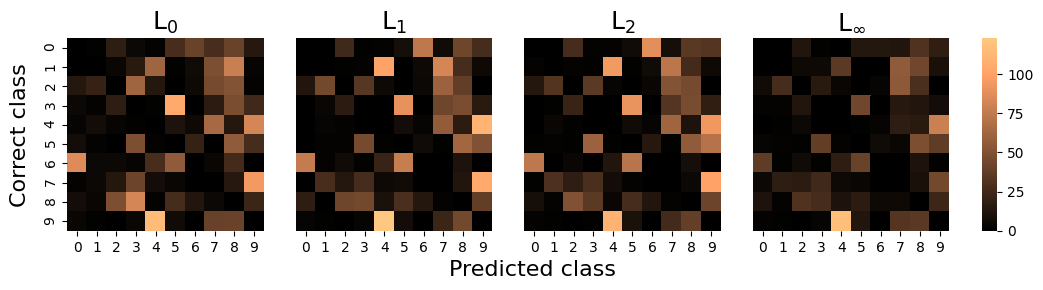}
\vspace{-0.75em}
\caption{Visualisation of the number of successful attacks on the convolutional network trained on MNIST, using four attack methods.}
\label{fig:confusion_matrix}
\end{figure*}

To provide statistically sound results, we choose original (correct) and predicted (incorrect) class (i.e. $o,p$) according to the confusion matrix (see Fig.~\ref{fig:confusion_matrix}) of the generated AEs. We pick such combinations ($o,p$), which have a lot of occurrences ($\approx$100) in the table. To compare different attack methods we can fix the pair ($o,p$) thanks to the fact that distinct attacks tend to have similar success rate for all possible pairs (also seen from Fig.~\ref{fig:confusion_matrix}).

\begin{figure*}[t!]
\centering
\includegraphics[width=0.255\columnwidth]{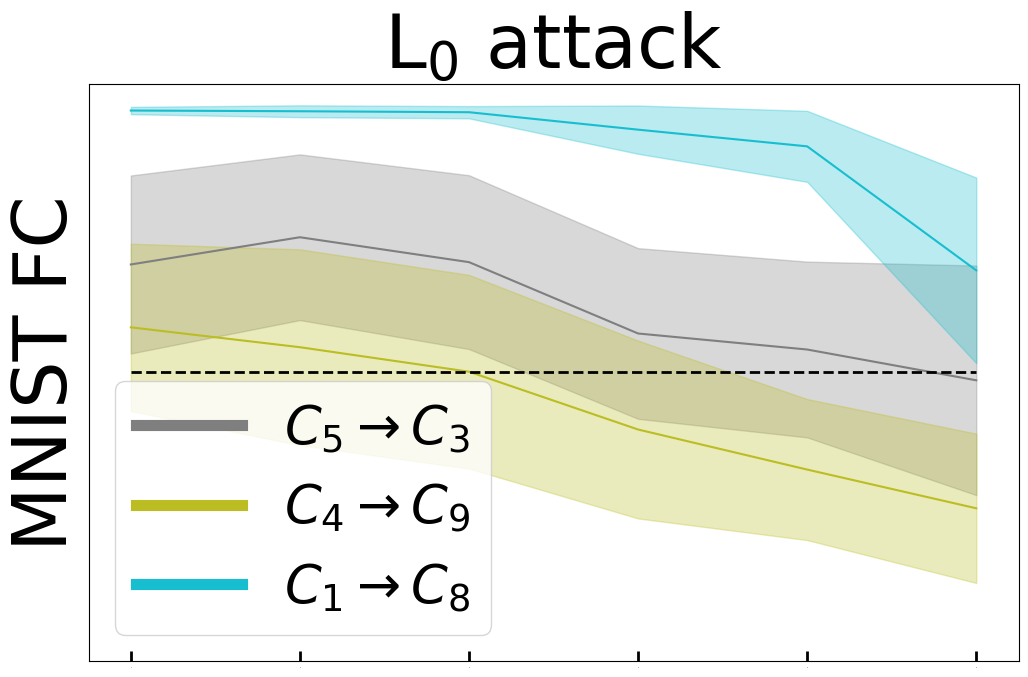}
\includegraphics[width=0.238\columnwidth]{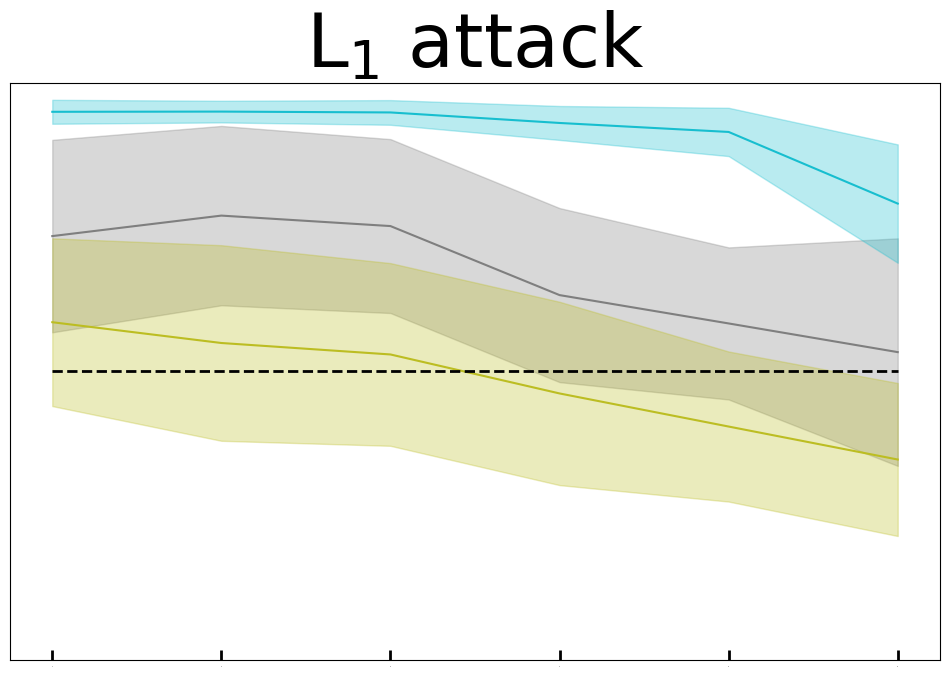}
\includegraphics[width=0.238\columnwidth]{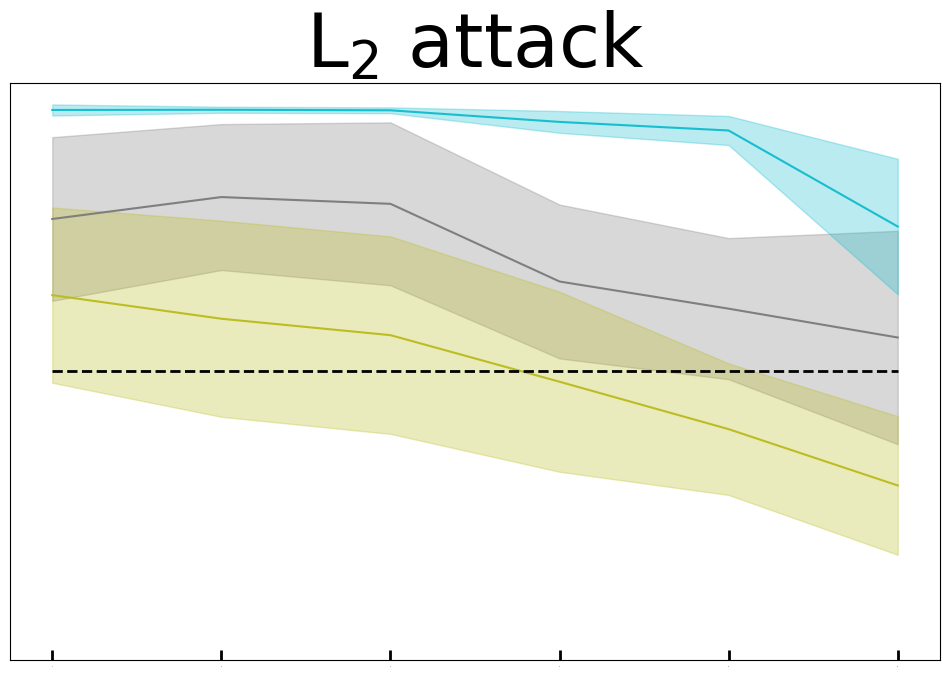}
\includegraphics[width=0.238\columnwidth]{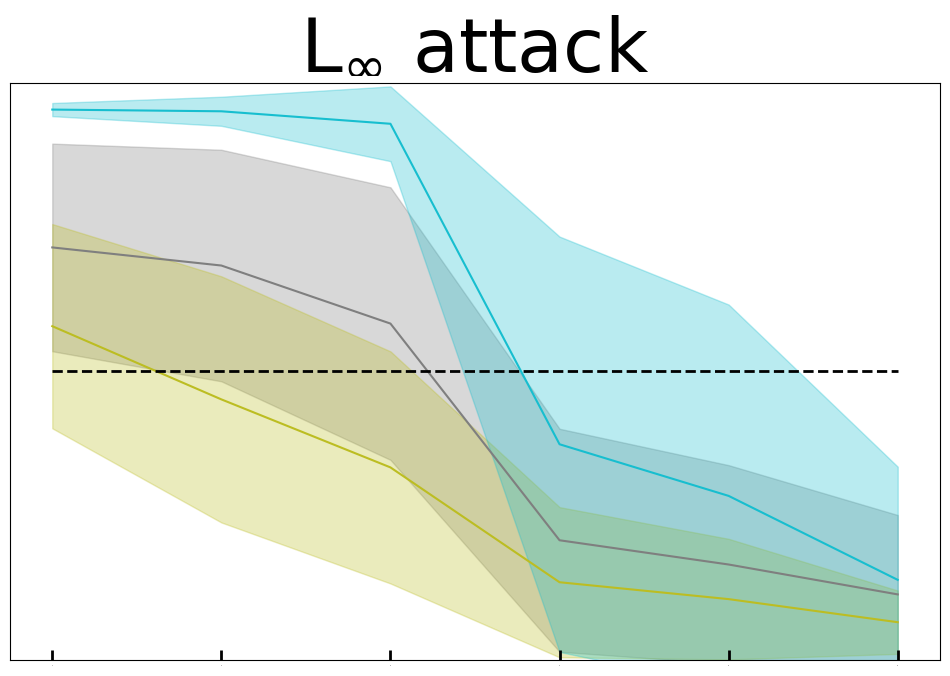}
\includegraphics[width=0.255\columnwidth]{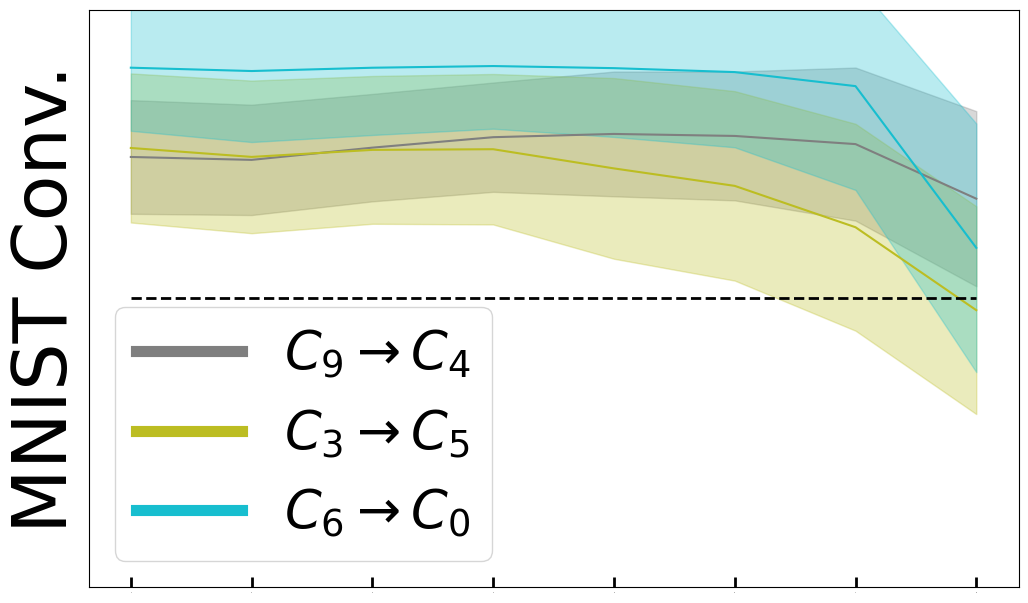}
\includegraphics[width=0.238\columnwidth]{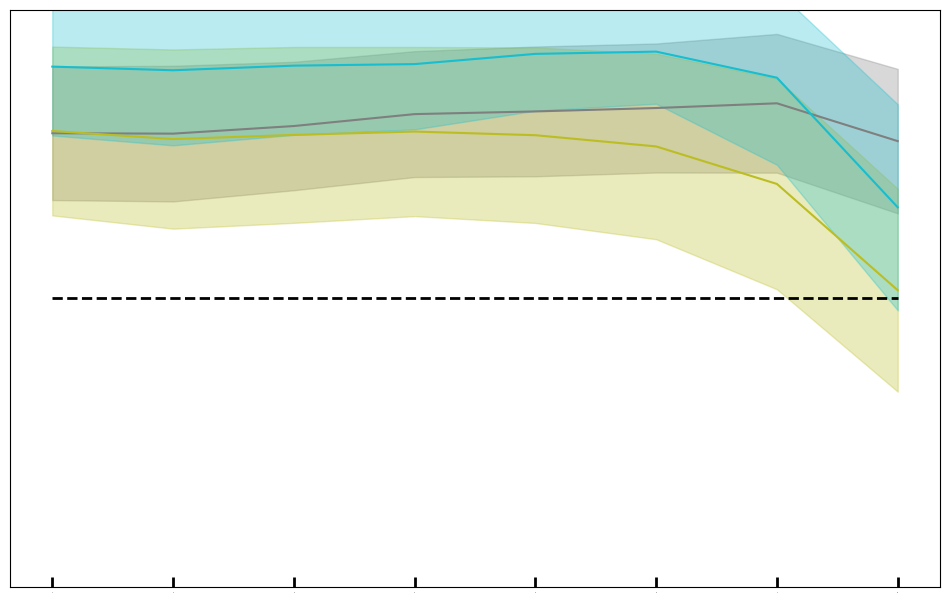}
\includegraphics[width=0.238\columnwidth]{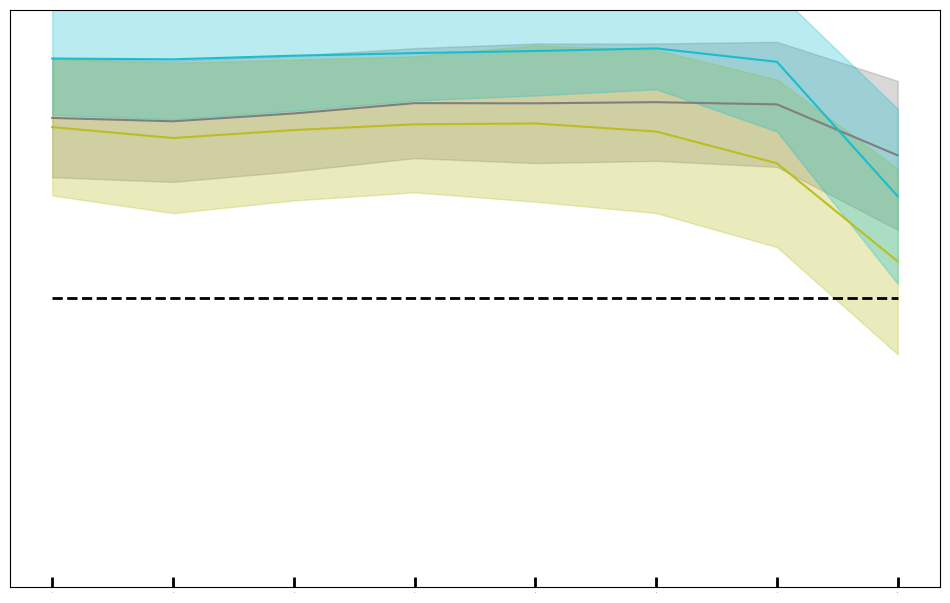}
\includegraphics[width=0.238\columnwidth]{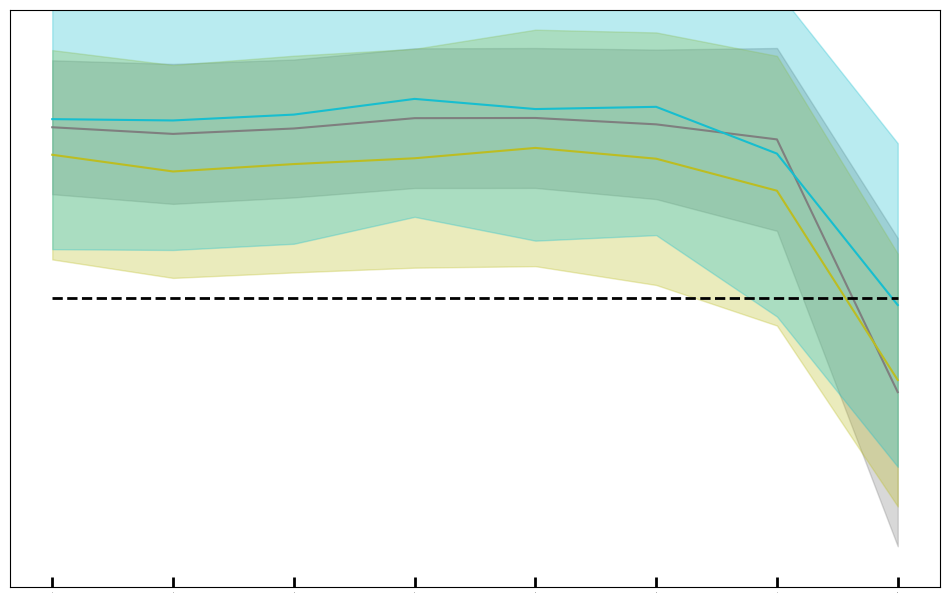}
\includegraphics[width=0.255\columnwidth]{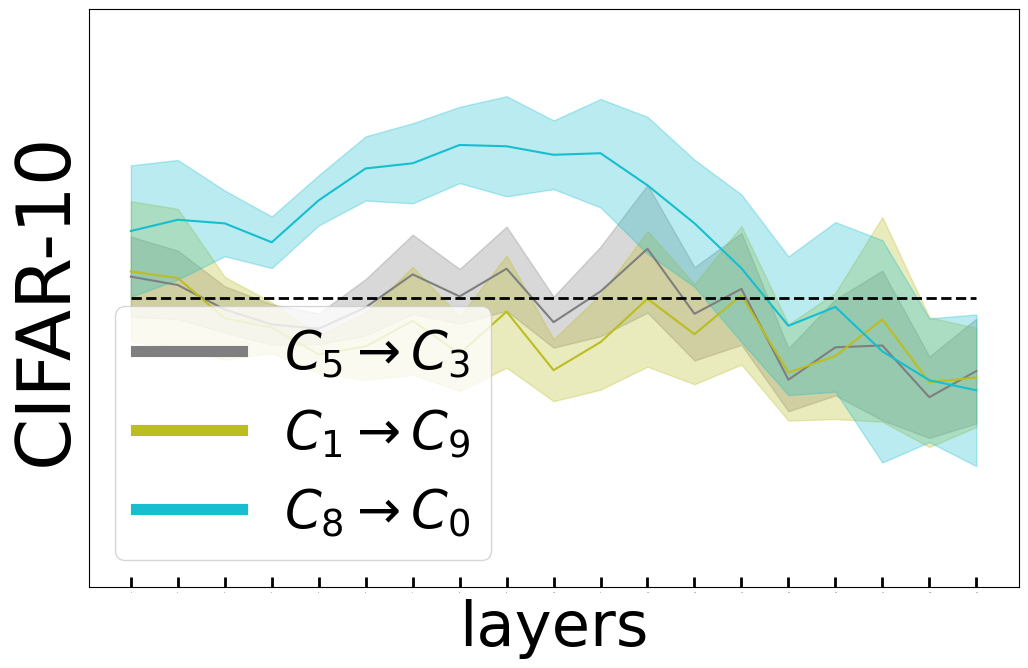}
\includegraphics[width=0.238\columnwidth]{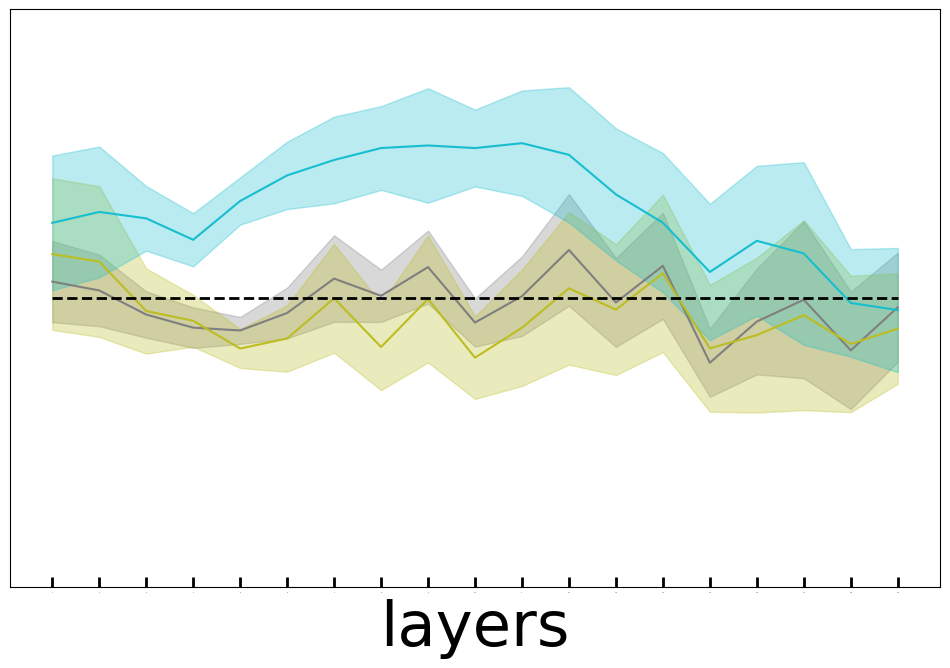}
\includegraphics[width=0.238\columnwidth]{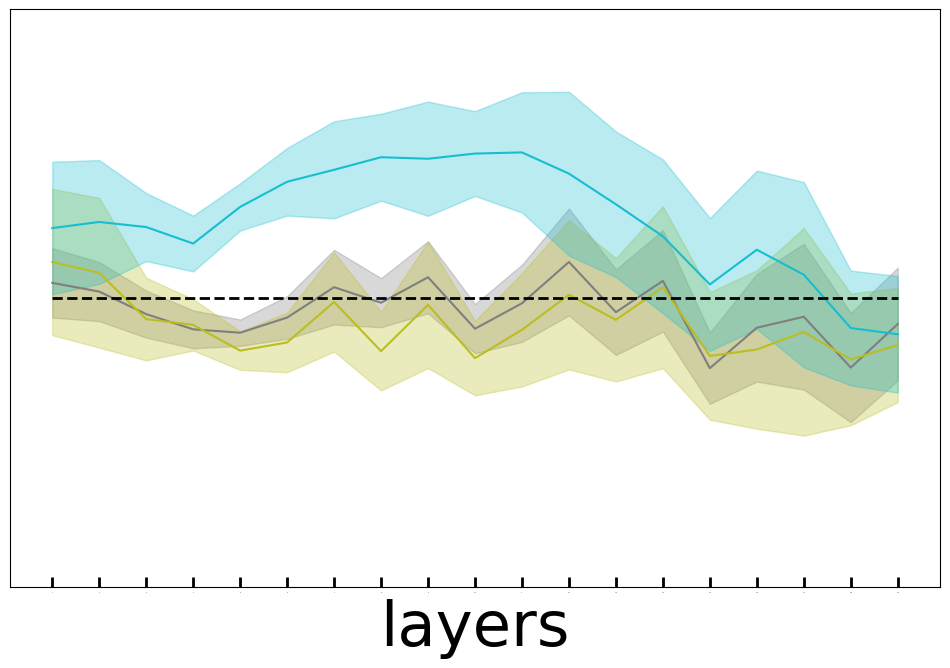}
\includegraphics[width=0.238\columnwidth]{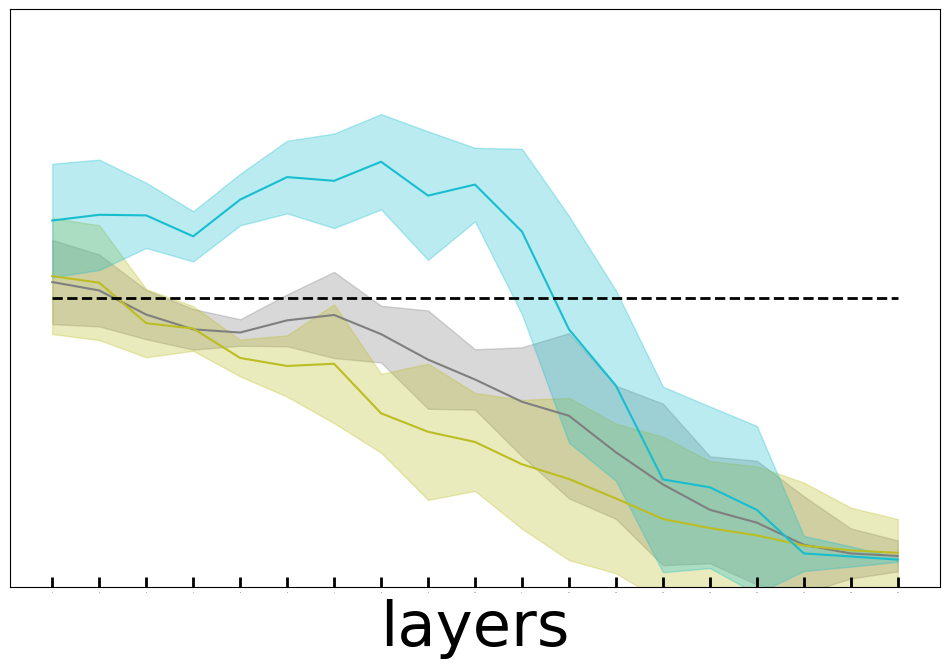}
\caption{Development of the $k_o/k$ ratio ($y$-axis) along the networks ($x$-axis) for three chosen pairs of ($o$, $k$) for each network, using different attacks. The dashed horizontal line in the middle of the plot helps to visualise, where the number of neighbours to the correct class is lower than those to the predicted class.}
\label{fig:dist_ratio}
\end{figure*}

Results in Fig.~\ref{fig:dist_ratio} illustrate that in the first layers the AEs usually have a lot of neighbours belonging to the correct class. 
Surprisingly, this often holds for the successive layers as well. Individual attacks have their own behaviour, regarding the distance to the manifolds. AEs crafted using $L_0$ tend to eventually have more neighbours from the incorrect class, however only by a small margin. $L_1$ and $L_2$ attacks show very similar behaviour, since these AEs often have more neighbours from the correct class, even near the last hidden layer. $L_{\infty}$ attack invokes the strongest reaction, since the AEs have the tendency to completely flip the number of neighbours, in case of the fully connected network and the network trained on CIFAR-10, after several layers almost all neighbours are from the incorrect class. For the convolutional network trained on MNIST this flip is only visible at the end. 
We also observe that each of the explored pairs ($o,p$) has its unique behaviour but follow a certain trend.

\subsection{Computing distances to class-specific manifolds} 

This method is based on the idea of measuring the distances between AE activations at hidden layers and class-specific manifolds (of the classes $C_o$ and $C_p$) of the train-set activations, so we apply it to a fixed attack and a subset of AEs ($Adv_{C_o \rightarrow C_p}$). A class-specific manifold is approximated by the convex hull of $k$ nearest neighbours (of a current AE activation) belonging to the given class. The distance to this manifold is the Euclidean distance to the (orthogonal) projection on the convex hull. The projection is easily expressed as a constrained convex minimisation problem
\begin{align}
\begin{split}
    &\text{minimize}_{\alpha_1,\alpha_2,\dots,\alpha_k} \quad \Bigg\| \left(\sum_{i=1}^k\alpha_i\mathbf{x}_i\right)-\mathbf{x}\Bigg\|_2, \\
    &\text{subject to} \quad \sum_{i=1}^k\alpha_i=1,\quad \alpha_i\geq0,\ i \in \{1, 2,\dots,k\},
\end{split}
\label{eq:conv_projection}
\end{align}
where $\mathbf{x}_i$ are the nearest neighbours of $\mathbf{x}$ approximating the manifold.

\begin{figure*}[t!]
\centering
\includegraphics[width=0.425\columnwidth]{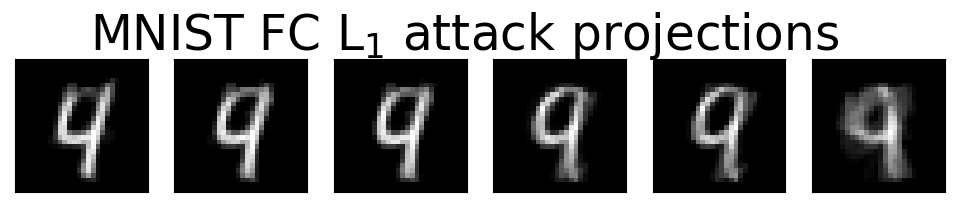}
\includegraphics[width=0.567\columnwidth]{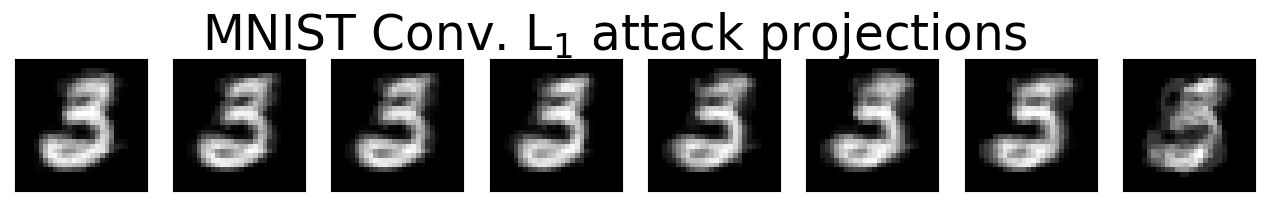}
\includegraphics[width=0.425\columnwidth]{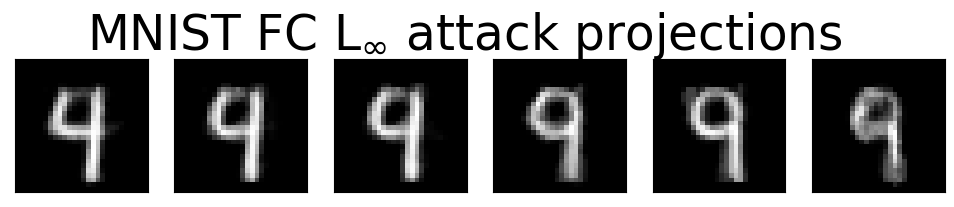}
\includegraphics[width=0.567\columnwidth]{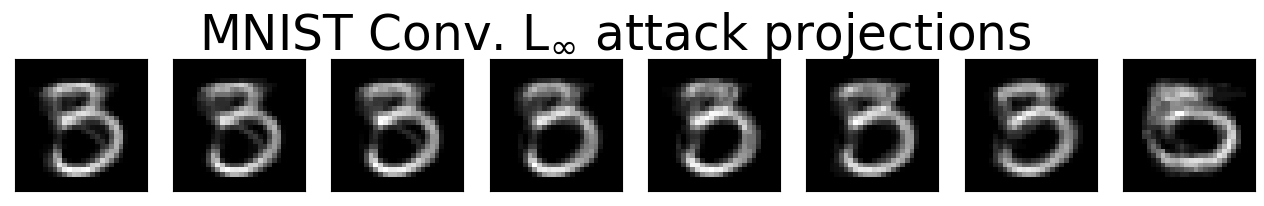}
\caption{Convex combinations of inputs computed using coefficients from projections of activations of AEs in the cases of $Adv_{C_4 \rightarrow C_9}$ (fully-connected network) and $Adv_{C_3 \rightarrow C_5}$ (convolutional network).}
\label{fig:projections}
\end{figure*}

\begin{figure*}[t!]
\centering
\includegraphics[width=0.262\columnwidth]{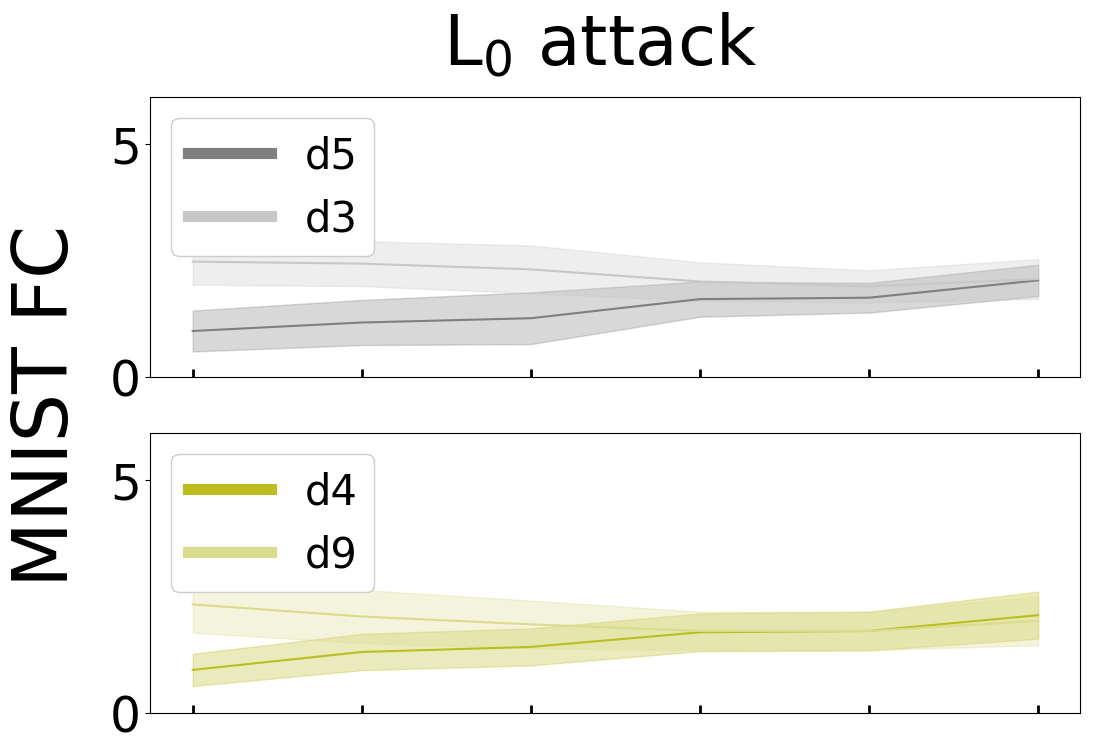}
\includegraphics[width=0.236\columnwidth]{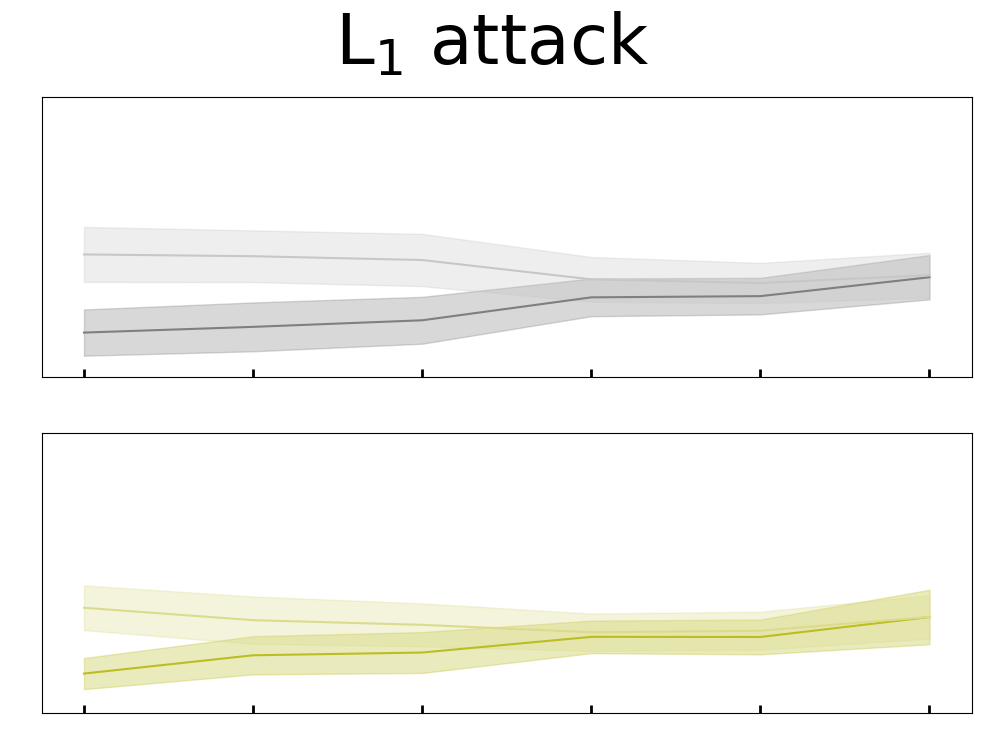}
\includegraphics[width=0.236\columnwidth]{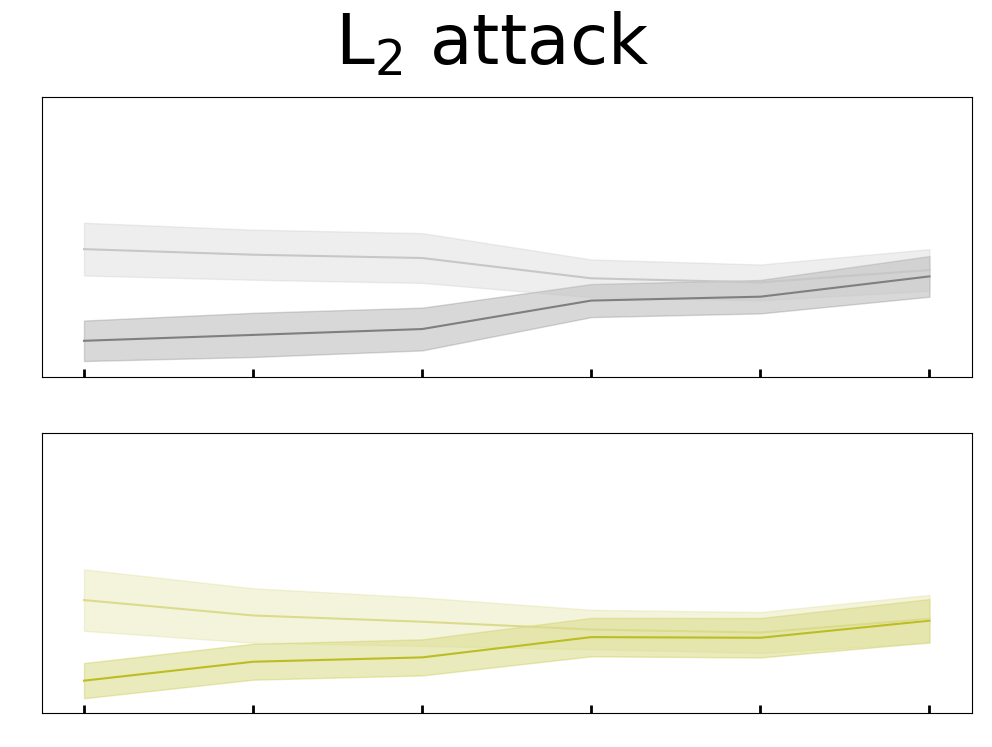}
\includegraphics[width=0.236\columnwidth]{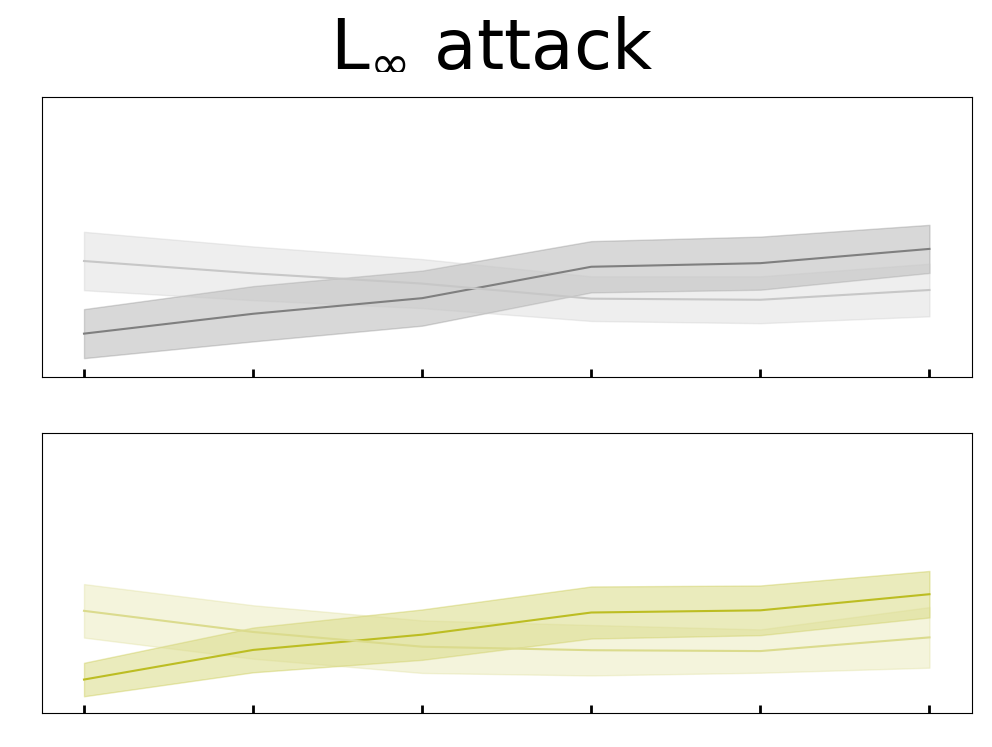}
\includegraphics[width=0.262\columnwidth]{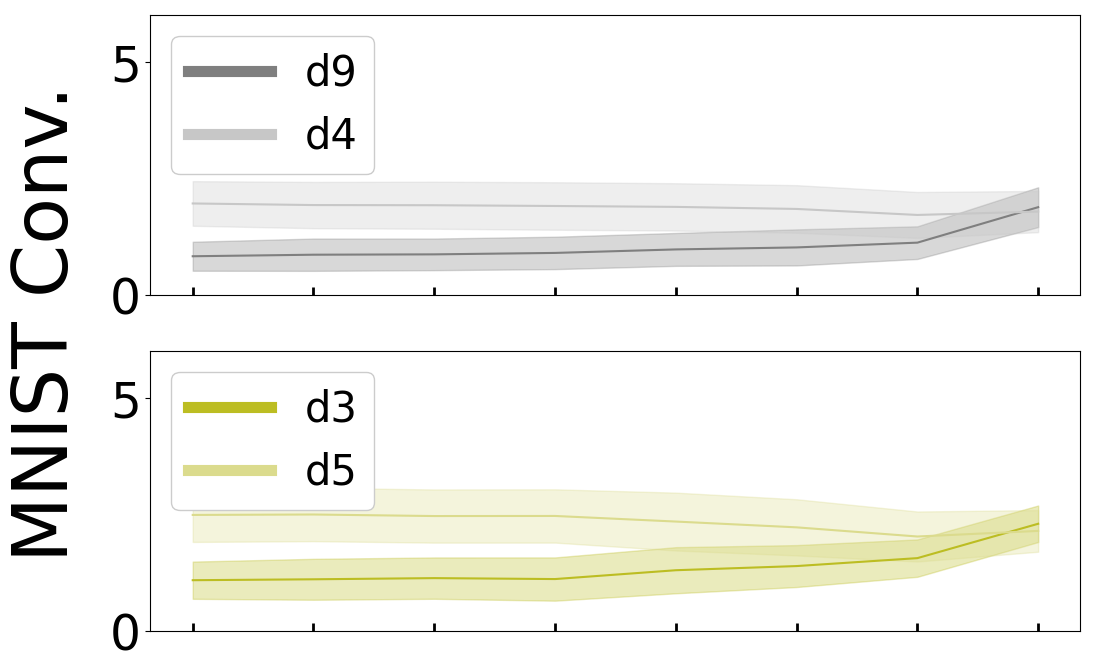}
\includegraphics[width=0.236\columnwidth]{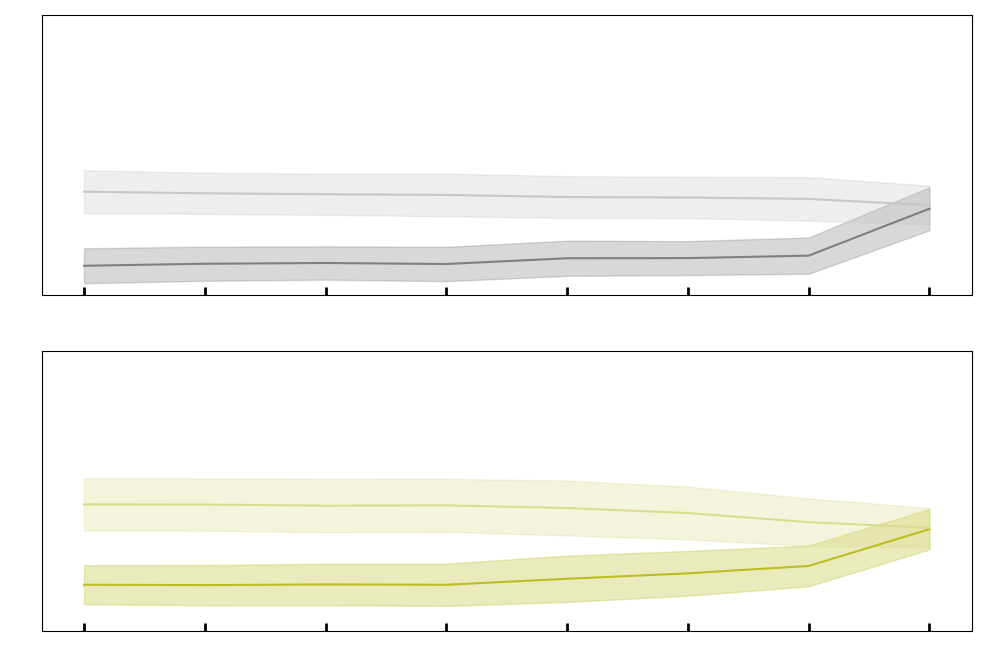}
\includegraphics[width=0.236\columnwidth]{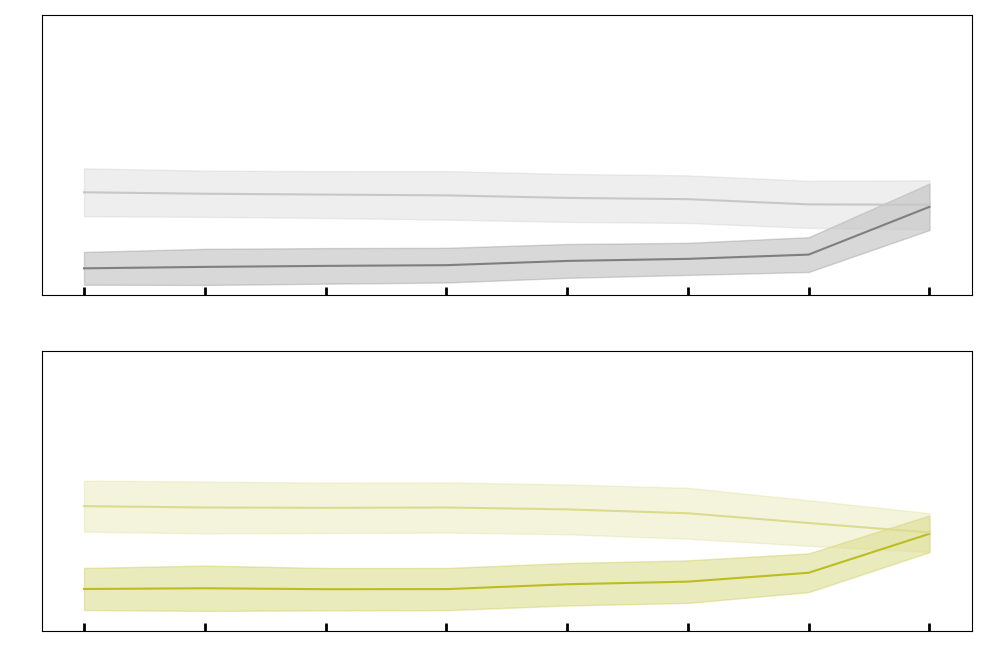}
\includegraphics[width=0.236\columnwidth]{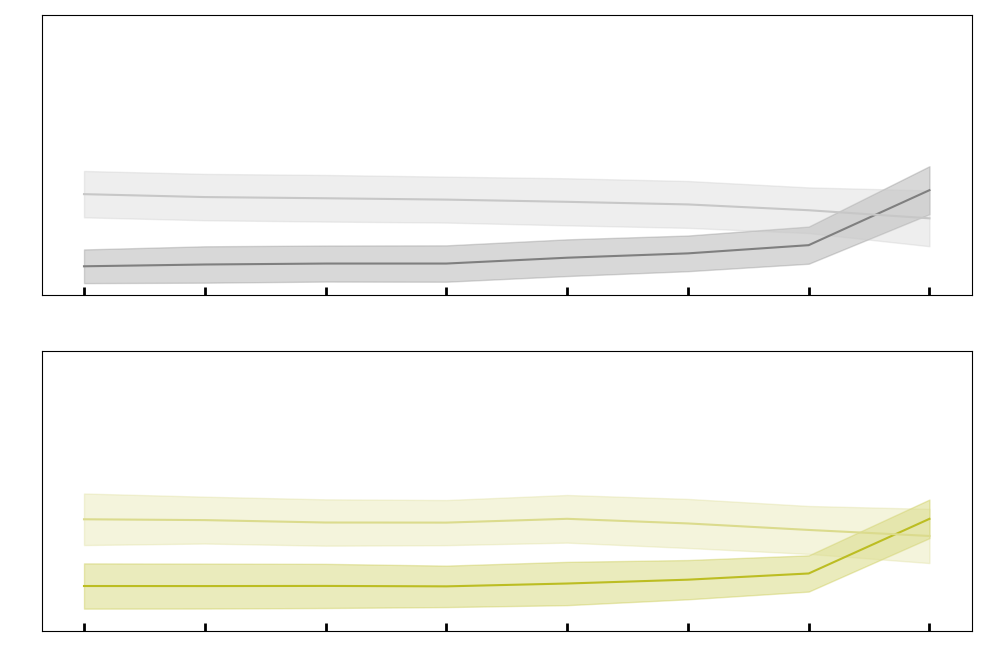}
\includegraphics[width=0.262\columnwidth]{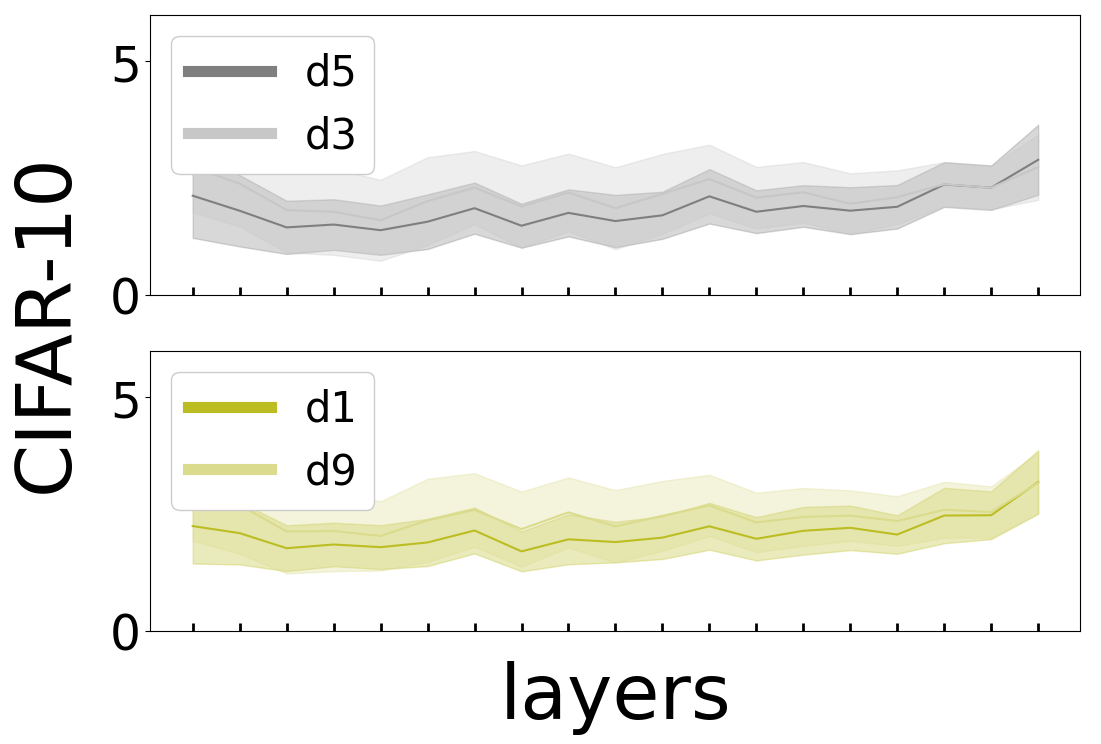}
\includegraphics[width=0.236\columnwidth]{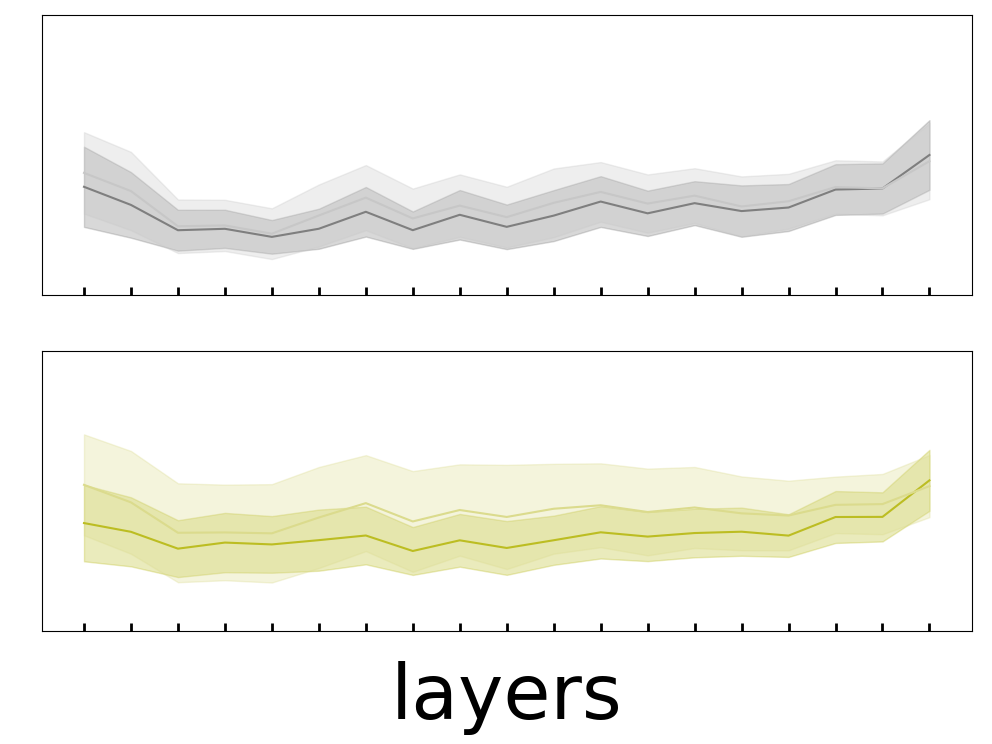}
\includegraphics[width=0.236\columnwidth]{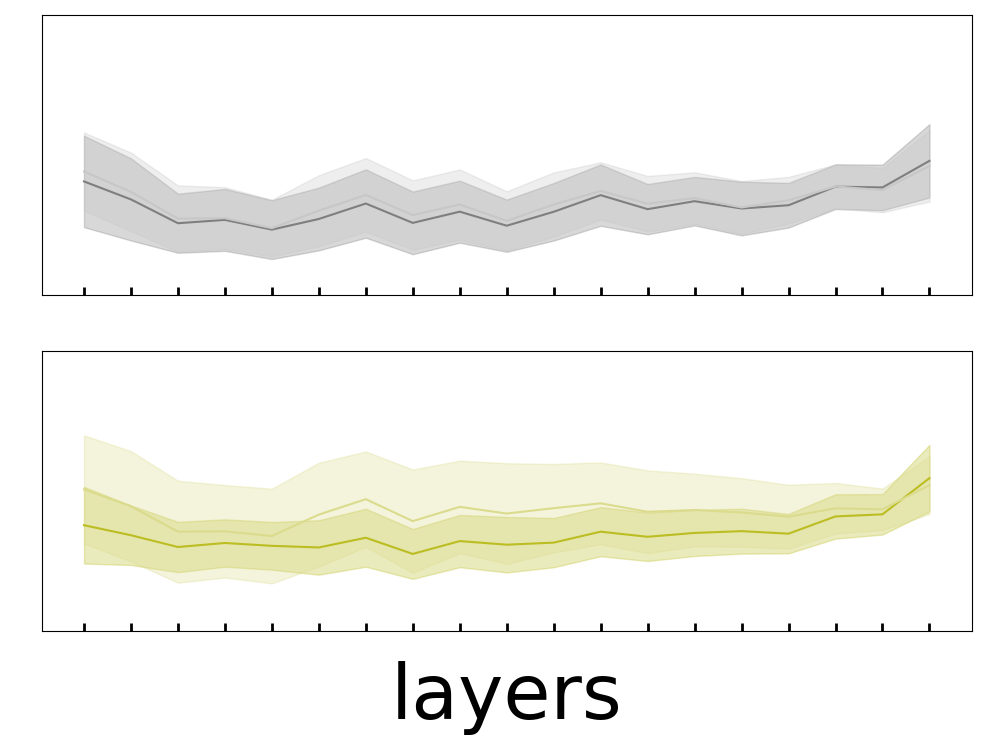}
\includegraphics[width=0.236\columnwidth]{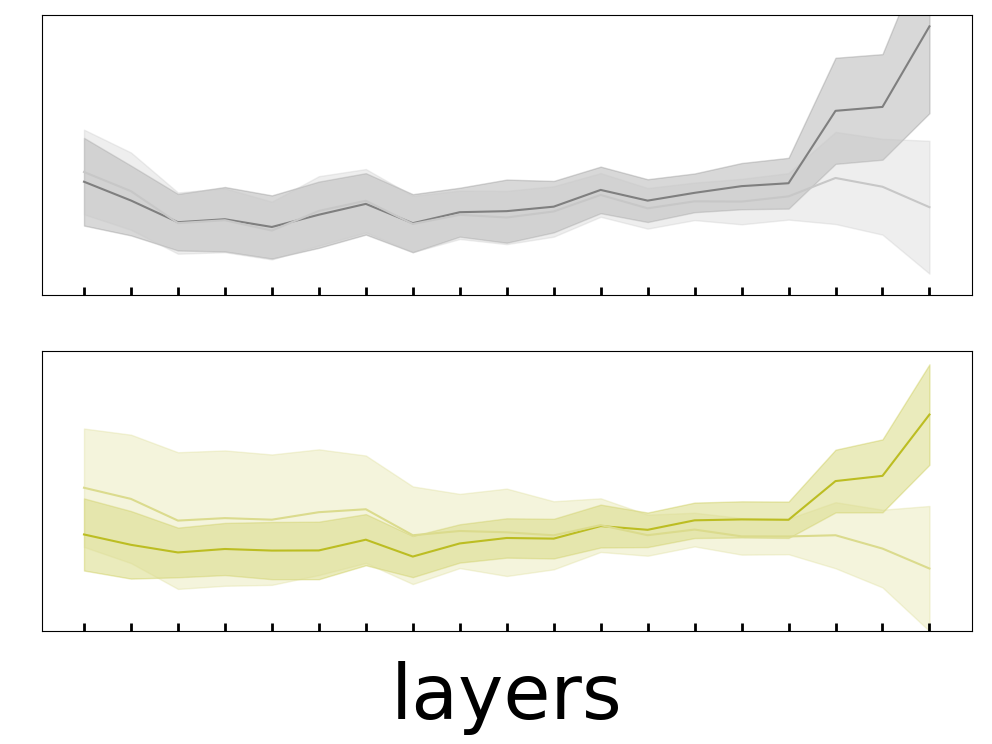}
\caption{Development of the distances ($y$-axis) of convex combinations of input images, corresponding to AEs activation projections, to the class-specific manifolds of $C_o$ and $C_p$. The $x$-axis denotes individual layers. In general, we observe a trend of AEs activations moving away from the original class manifold towards the predicted class manifold.}
\label{fig:dist_projected}
\end{figure*}

However, the dimensionality of hidden layers varies so the comparison of distances between different layers would be inconsistent. In order to avoid this problem, we proceed as follows: for each layer and each AE activation, we first compute the projection onto the manifold of the entire train-set activations (i.e., we do not require them to belong to a specific class) but instead of computing the distance to this projection, we only remember the indices of the $k$-NN and their respective coefficients $\alpha_i$ (Eq. \ref{eq:conv_projection}). These are used to compute the corresponding convex combination of images in the input space (a few examples of such projections across network layers are shown in Fig. \ref{fig:projections}), which is then projected onto the class-specific manifolds of inputs belonging to classes $C_o$ and $C_p$. Then we can reliably monitor the development of these distances across the network. Results are shown in Fig.~\ref{fig:dist_projected}. Both chosen pairs ($o,p$) were also used in the first method, allowing a better comparison.
The results are quite consistent, PGD attack has the greatest influence on the distances to the class manifolds. In almost all cases, the distance to the predicted class decreases through the network layers. In the fully-connected MNIST network, the change is gradual, in the other two networks, it happens mostly on the last few hidden layers.

\section{Assessment of entanglement}

Manifold disentanglement theory states that the manifolds of individual classes tend to separate (disentangle) during the classification, although they are entangled in the earlier layers \cite{Brahma2016}. In this section we shed light on what happens during manifold disentanglement with adversarial examples applying two methods: First, we visualise the projections of manifolds using the state-of-the-art nonlinear dimensionality reduction method UMAP \cite{McInnes2018} and second, we numerically determine the exact ``rates'' of AEs dis/entanglement. We focus on the differences between the representations of the AEs and ordinary test set examples.

\begin{figure}[ht!]
  \begin{minipage}{0.91\columnwidth}
    \begin{minipage}{0.455\columnwidth}
      \includegraphics[width=\columnwidth]{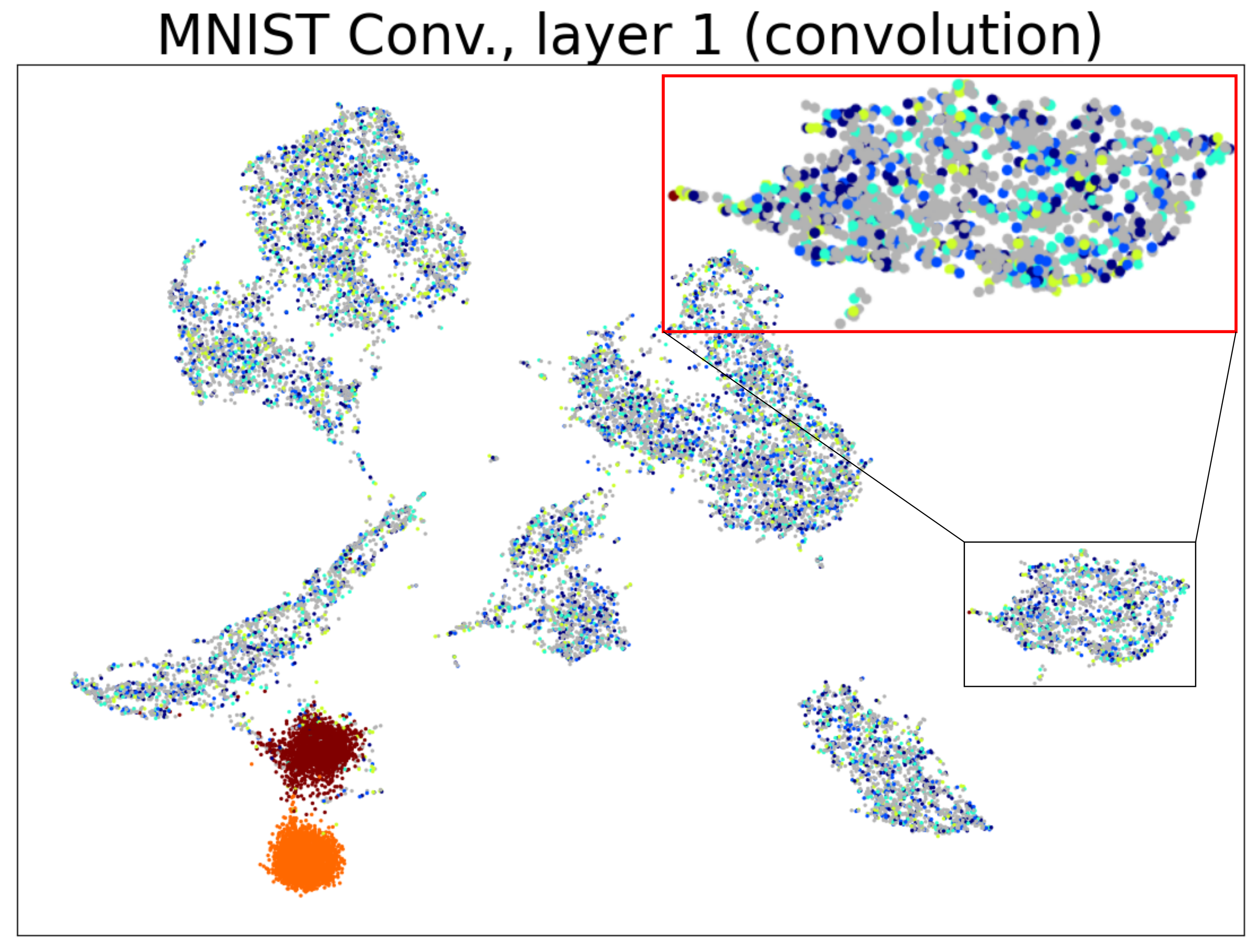}
    \end{minipage} 
    \hspace{\columnsep}
    \begin{minipage}{0.455\columnwidth}
      \includegraphics[width=\columnwidth]{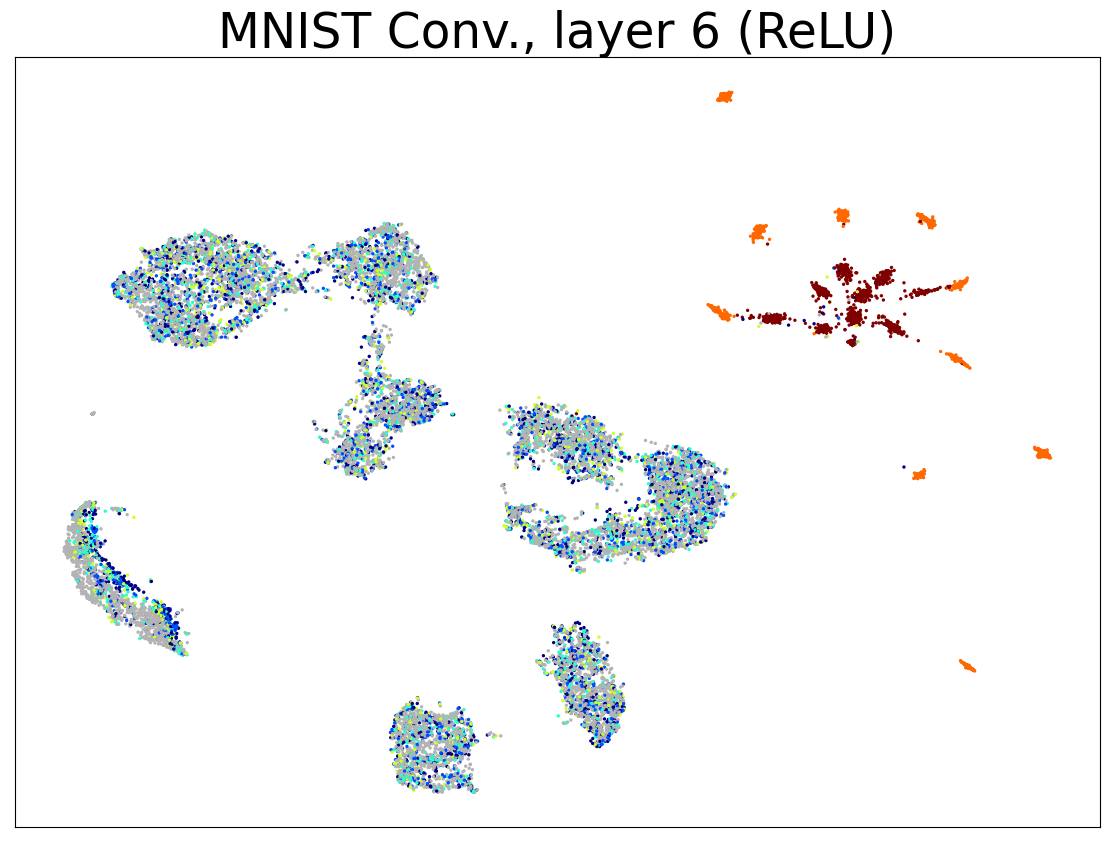}
    \end{minipage} 
    \\
    \begin{minipage}{0.455\columnwidth}
      \includegraphics[width=\columnwidth]{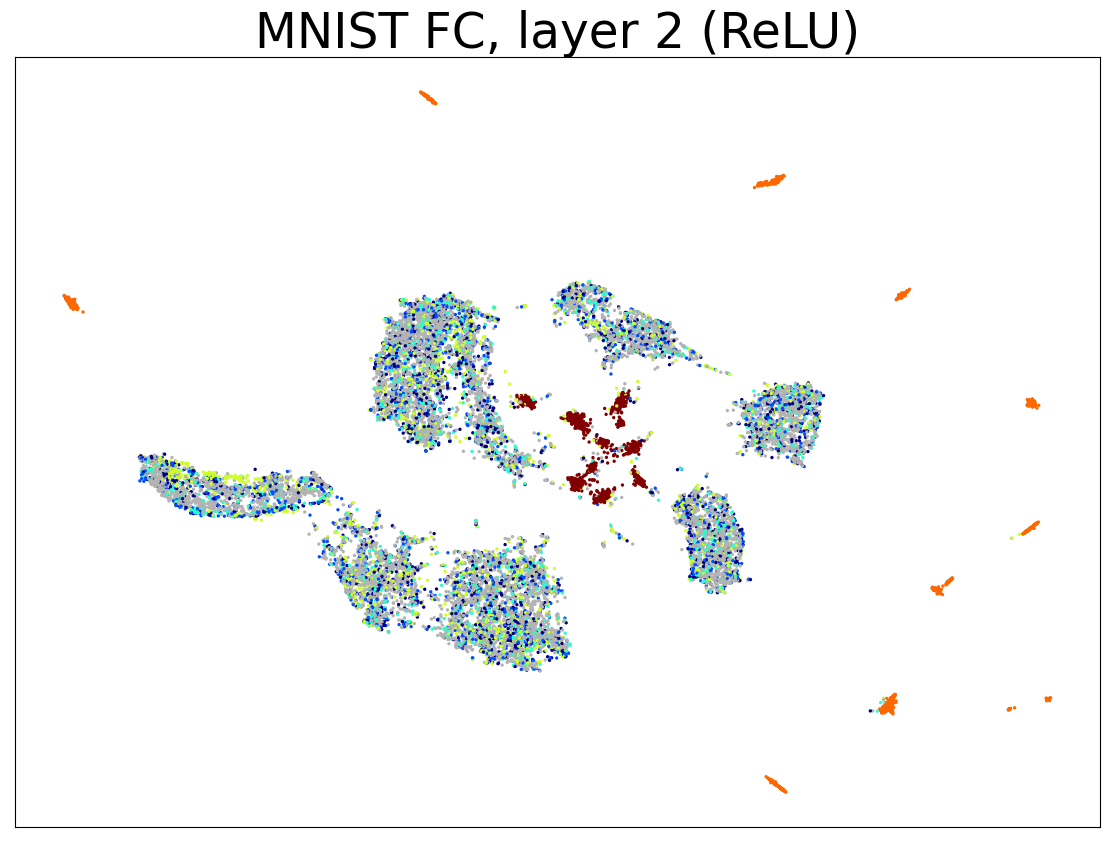}
    \end{minipage} 
    \hspace{\columnsep}
    \begin{minipage}{0.455\columnwidth}
      \includegraphics[width=\columnwidth]{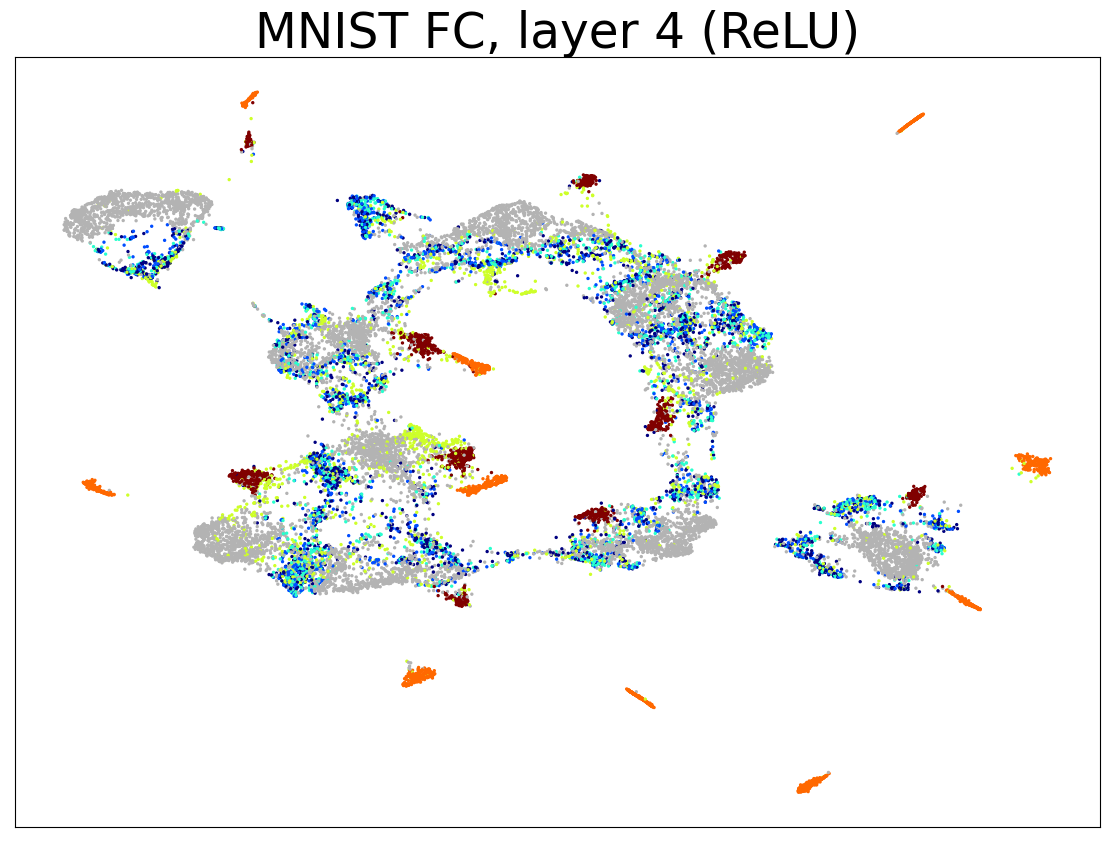}
    \end{minipage} 
  \end{minipage}
  \hspace{-0.5em}
  \begin{minipage}{0.09\columnwidth}
    \includegraphics[width=\columnwidth]{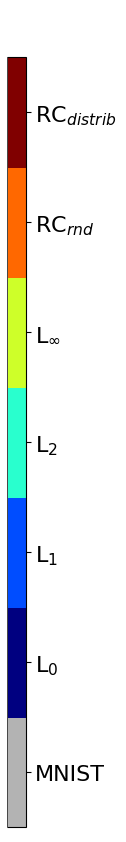}
  \end{minipage}
  \caption{Visualisation of  activations of four types of adversarial examples, two types of rubbish class examples and the test set, using UMAP. Distinctive behaviours depend on the network and the layer: very high entanglement of AEs with the test set (top left) is contrasted with a minimum entanglement (bottom right).}
  \label{fig:umap_conv}
\end{figure}

Our experiment proceeds as follows: we start with the whole set of AEs together with the test set, and feed them into the trained network. Usually, to classify an image, one only needs to care about the last layer, but here we are interested in activations on each hidden layer. Next, a dimensionality of representations on each hidden layer is (nonlinearly) reduced into two dimensions using UMAP. For the final step, the resulting points are depicted in 2D space, where each type of adversarial has its unique color, in order to highlight their distributions. The output of the mentioned method can be seen in Fig.~\ref{fig:umap_conv}.
We observe the following phenomena: 
\vspace{-0.5em}
\begin{enumerate}
\item On most of the layers, all 4 types of AEs are close to the activations of the test set, one could say they are entangled. This observation is valid for all three networks. 
\item The only visual distinction between the clusters of AEs and the test set can be seen towards the last layers. The attack with $L_{\infty}$ constraint shows the largest deviation. 
\item Rubbish class examples are well separated on all layers, where usually ${\rm RC}_{rnd}$ is a bit further from the rest of the data than ${\rm RC}_{distrib}$. Towards the last layers, both rubbish class sets are progressively split into ten clusters, corresponding to classes. 
This indicates a certain specificity for each class even in the seemingly random patterns.
\end{enumerate}

To verify whether the first two statements also hold for high-dimensional representations on the hidden layers, we measure the entanglement of adversarial vs. non-adversarial inputs. The standard methods to do so are the triplet loss \cite{triplet}, the soft nearest neighbour loss \cite{frosst_nll} or the comparison of interclass/intraclass distances. For our analysis, we employ the soft nearest neighbour (SNN) loss defined as
\begin{equation}
    l_{\rm SNN}(\mathbf{X},\mathbf{y},T) = - \frac{1}{b} \sum\limits_{i \in \{1..b\}}\log \left( \frac{\sum\limits_{\substack{j \in \{1..b\}\setminus i  \\ y_i = y_j}}\exp(-\|\mathbf{x}_i - \mathbf{x}_j\|^2/T)}{\sum\limits_{\substack{k \in \{1..b\}\setminus i }}\exp(-{\|\mathbf{x}_i - \mathbf{x}_k\|^2/T)}} \right)
\label{eq:softnnl}
\end{equation}
where $\mathbf{X}$ are the input points (as a matrix), $\mathbf{y}$ are the corresponding categories, $b$ is the batch size and $T$ is the temperature. To reduce the number of parameters, we perform SGD over $T$ (minimizing the SNN loss), following \cite{frosst_nll}. Due to the large  distances in high-dimensional spaces, the optimization can sometimes be unstable. Therefore we did not use the SNN loss to analyse the CIFAR-10 network. 

\begin{figure*}[t!]
\centering
\includegraphics[width=0.49\columnwidth]{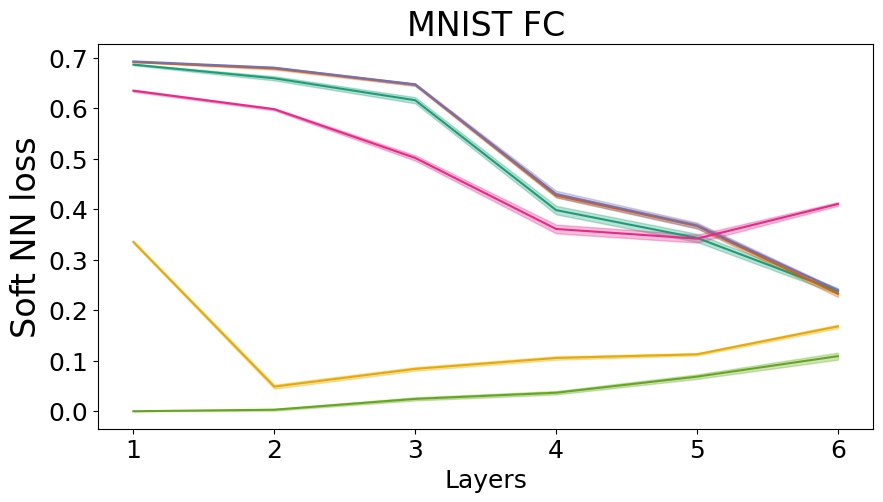}
\includegraphics[width=0.465\columnwidth]{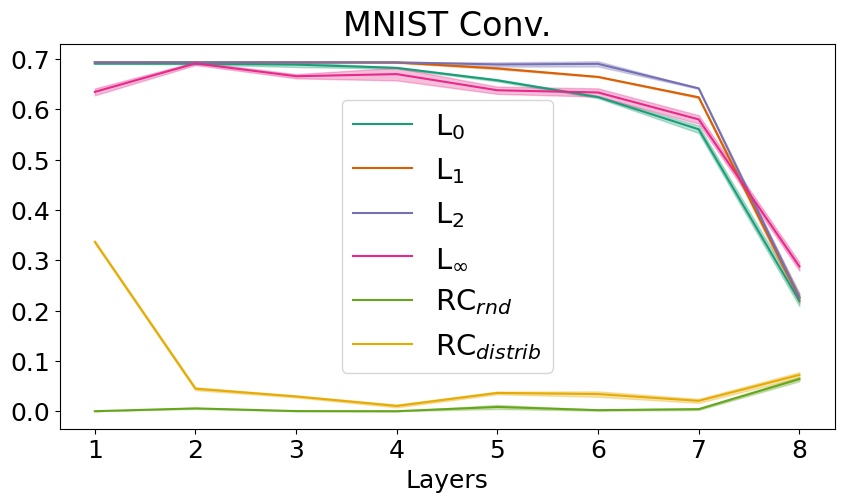}
\caption{Visualisation of the soft nearest neighbour loss scores between malicious inputs (AEs and the rubbish class) and the test set (averaged across 4 runs) throughout the networks trained on MNIST.}
\label{fig:soft_nn_loss}
\end{figure*}

The SNN loss assigns high values to entangled data, whereas low values suggest that the data is disentangled, usually forming nice and separate clusters. By calculating the SNN loss, we found that the activations of AEs are indeed entangled with the activations of the test set (see Fig.~\ref{fig:soft_nn_loss}). 
The rubbish class examples showed quite the opposite behaviour by being almost completely disentangled from the test set. Out of the four types of AE-generating methods, PGD leads to faster disentanglement in the network, the others start to disentangle only towards the final layer. We also see big differences between the individual types of networks. In the fully-connected network the process of disentangling is gradual, whereas in the convolutional network the disentangling is only visible on the last few layers.

\section{Conclusion}

We studied the behaviour of adversarial examples on the hidden layers of neural networks trained on MNIST and CIFAR-10 datasets. We confirmed that AEs generated using different $L_p$ norm constraints invoke various effects in the neural network, thus the cause of the misclassification can also differ. 

Following a simple statistical analysis of distances of AEs to different classes, we proposed two novel methods of analysing these distances on hidden layers, allowing for a more consistent comparison throughout the studied networks. The first method is based on the deep $k$-NN, where we count the number of nearest neighbours only from specific classes and monitor their ratio. The second method also utilizes the nearest neighbours, but they are used to approximate the manifold, onto which the AEs are projected. After the projection is calculated, we perform the distance measurement in the input space, thus alleviating the problem of comparison of distances with spaces of different dimensions. Both methods showed consistent results that some of the AEs (mainly those generated using PGD) tend to approach the manifold of the predicted (incorrect) class faster in the network. In the other cases, the misclassification often occurs even though the activations of AEs are closer to the manifold of the correct class even in the last layers. 

Our analysis using the SNN loss reveals that AEs are entangled with the test set. This is the case of hidden layer representations as well, where AEs show a high level of entanglement which slightly decreases towards the last layers of the studied networks. This does not hold for the rubbish class examples, where the opposite is true. Those examples begin disentangled in the input space, and gradually get a little entangled with the test set. However, they do not get entangled too much before they reach the last layer. We also confirmed the entanglement visually, using UMAP, where after the transformation of the activation space into 2D we see, that AEs co-occupy the areas with the test set examples. 

To extend our work, it would be interesting to see how the measured behaviour changes using adversarially trained networks and whether it is possible for AEs to have similar behaviour as in adversarially trained ones, without performing the adversarial training. Our work may serve as an inspiration for evaluating novel attacks, or as a score to pick the right methods of adversarial training.

\section*{Acknowledgments}
This research was partially supported by TAILOR, a project funded by EU Horizon 2020 research and innovation programme under GA No 952215.

\bibliographystyle{unsrt}  
\bibliography{references}

\end{document}